\documentclass[10pt,journal]{IEEEtran}
\usepackage{graphicx,subcaption}
\usepackage{amsmath}
\usepackage{bm}
\usepackage{amssymb}
\usepackage{color, colortbl}
\usepackage{diagbox}
\usepackage{multirow}
\usepackage{url}
\usepackage[nocompress]{cite}
\usepackage{arydshln}
\usepackage{hyperref}
\usepackage{xcolor}

\hypersetup{
    colorlinks,
    linkcolor={red!70!black},
    citecolor={blue!70!black},
    urlcolor={blue!70!black}
}

\ifCLASSOPTIONcompsoc
  \usepackage[nocompress]{cite}
\else
  \usepackage{cite}
\fi

\ifCLASSINFOpdf
\else
\fi

\begin{document}

\title{What's color got to do with it?  Face recognition in grayscale}

\author{$^1$Aman~Bhatta,
        $^2 {^\dagger}$Domingo~Mery\thanks{$\dagger$ Dr. Mery thanks the National Center for Artificial Intelligence CENIA FB210017, Basal ANID, Chile.},~\IEEEmembership{Member,~IEEE},
        $^1$Haiyu~Wu, $^3$Joyce Annan, 
        $^3$Michael C. King,~\IEEEmembership{Senior Member,~IEEE} 
        and~$^1 {^\ddagger}$Kevin~W.~Bowyer\thanks{$\ddagger$ Dr. Bowyer is a member of the FaceTec (\url{facetec.com}) Advisory Board.  Results in this paper do not necessarily relate to FaceTec products.},~\IEEEmembership{Fellow,~IEEE}\\~\\
        $^1$University of Notre Dame, Notre Dame, Indiana \\
        $^2$Pontificia Universidad Católica de Chile, Macul, Chile \\
        $^3$Florida Insitute of Technology, Melbourne, Florida

}%

\markboth{Preprint submitted for Review.}%
{Shell \MakeLowercase{\textit{et al.}}: Bare Advanced Demo of IEEEtran.cls for IEEE Biometrics Council Journals}

\IEEEtitleabstractindextext{%
\begin{abstract}

State-of-the-art deep CNN face matchers are typically created using extensive training sets of color face images. Our study reveals that such matchers attain virtually identical accuracy when trained on either grayscale or color versions of the training set, even when the evaluation is done using color test images. Furthermore, we demonstrate that shallower models, lacking the capacity to model complex representations, rely more heavily on low-level features such as those associated with color. As a result, they display diminished accuracy when trained with grayscale images. We then consider possible causes for deeper CNN face matchers ``not seeing color''. 
Popular web-scraped face datasets actually have 30 to 60\% of their identities with one or more grayscale images.
We analyze whether this grayscale element in the training set impacts the accuracy achieved, and conclude that it does not.
We demonstrate that using only grayscale images for both training and testing achieves accuracy comparable to that achieved using only color images for deeper models.
This holds true for both real and synthetic training datasets.
HSV color space, which separates chroma and luma information, does not improve the network's learning about color any more than in the RGB color space. 
We then show that the skin region of an individual's images in a web-scraped training set exhibits significant variation in their mapping to 
color space. This suggests that color
carries limited identity-specific information.
We also show that when the first convolution layer is restricted to a single filter, models learn a grayscale conversion filter and  pass a grayscale version of the input color image to the next layer. Finally, we demonstrate that leveraging the lower per-image storage for grayscale to increase the number of images in the training set can improve accuracy of the face recognition model.

\end{abstract}

}

\maketitle

\IEEEdisplaynontitleabstractindextext

\IEEEpeerreviewmaketitle

\ifCLASSOPTIONcompsoc
\IEEEraisesectionheading{\section{Introduction}\label{sec:introduction}}
\else
\section{Introduction}
\label{sec:introduction}
\fi

{\it Achromatopsia is a condition characterized by a partial or total absence of color vision. People with complete achromatopsia cannot perceive any colors; they see only black, white, and shades of gray.} \cite{Medline}

Web-scraped, in-the-wild face datasets were popularized by Labeled Faces in the Wild (LFW) \cite{lfw}.
Numerous web-scraped, in-the-wild face datasets have been introduced since LFW, increasing in size seemingly every year. 
While the MS1MV2 dataset \cite{guo_ECCV_2016} continues to be a widely-used, more accurate versions of matchers may be trained using newer, larger datasets such as Glint360K \cite{an_iccv_2021} and WebFace \cite{zhu_CVPR_2021}.  
All of these training sets contain images in RGB color format.
State-of-the-art face matchers such as ArcFace \cite{Deng_CVPR_2019}, Partial FC \cite{an_iccv_2021},  AdaFace \cite{Kim_CVPR_2022}, QMagFace \cite{terhorst_wacv_2023} and MagFace \cite{Meng_cvpr_2021}, use different loss functions in training a ResNet \cite{he_cvpr_2016} backbone, and the training sets for all of them contain images in RGB color format.
Color is essential for some general computer vision tasks \cite{Engilberge_ICIP_2017,de_2021_ICCV,singh_2020_arxiv}. 
But do current deep CNN face matchers actually use color to achieve better accuracy than they could using grayscale?

Building upon our previous work \cite{bhatta_wacvw_2024a}, this paper presents results showing that color does {\em not} result in higher accuracy face recognition than grayscale. In this comprehensive analysis,
\begin{itemize}
    \item We extend our previous results \cite{bhatta_wacvw_2024a} to include lighter and deeper backbones, and apply our findings to larger and more challenging datasets such as IJB-B and IJB-C. We also conduct multiple iterations of experiments to assess statistical significance of results.
    \item We expand our analysis to bottleneck the filters in the first layer to understand why models trained with grayscale perform well. By projecting the learned filter into RGB space, we show that essentially, the model trained with color first learns a grayscale conversion filter.
    \item We provide a more in-depth examination of how different images of the same identities in training datasets map onto color space.
    \item We explore whether this observation, which holds true for state-of-the-art (SoTA) face recognition models, also applies to a Commercial Off-The-Shelf (COTS) matcher.
    \item We investigate the effect of color within SoTA synthetic training datasets, employing multiple loss functions and varying backbone sizes.
\end{itemize}

This paper is organized as follows. Section \ref{litreview} gives a brief literature review. Sections \ref{pretrainedtest} and \ref{grayscaleimages}, together, analyze whether there is any accuracy difference between using grayscale or color images for the training data or the test data.  
Section \ref{implementation} details the network implementation for further experiments. Section \ref{colorspace} examines whether transforming into a colorspace that separates chroma and luma into different channels improves learning from color. Section \ref{deeperlook} looks at what the deep CNN learns in its early layer where the color image is input to the CNN. Section \ref{vartrainingdata} looks at how the skin region of different web-scraped, in-the-wild images of the same person maps to color space. Section \ref{filterforms} provides deeper analysis of the filters learned in the first layer by constraining the network to have fewer filters. Section \ref{synthetic} analyzes whether color is more relevant for synthetic training images. Section \ref{onechanneltraining} shows that training with a single-channel grayscale image leads to more efficient networks and could potentially also yield greater accuracy. Finally, Section \ref{conclusions} summarizes and discusses the results.

\section{Literature Review}\label{litreview}

\noindent\textbf{Impact of color on CNNs for general object classification/detection.} 
Researchers have investigated how noise, blur, jitter, compression, and other factors affect accuracy of general object classification by deep networks 
 \cite{dodge_2016_QoMEX,dodge_2017_ICCCN,ghosh_2018_ICASSP,zhou_2017_ICASSP,roy_2018_arxiv,pei_2019_tpami}. However, the impact of color has received less attention. One early study by Engilberge et al. \cite{Engilberge_ICIP_2017} analyzed the learned network to detect and characterize color-related features. They found color-specific units in CNNs and demonstrated that the depth of the layers affects color sensitivity. 
Buhrmester et al. \cite{buhrmester_2019_IbPRIA} investigated the impact of several color augmentation techniques on the deeper layers of the network and found that luminance is the most robust against changes in color systems. This finding suggests that the intensity value in color images contains the most useful content. De et al. \cite{de_2021_ICCV} showed that color information has significant impact on the inference of deep neural networks. Singh et al. \cite{singh_2020_arxiv} showed that CNNs often rely heavily on color information, but that this varies between datasets. Several researchers have found one color space better than another for general object classification \cite{diaz_2020_CVC, gowda_2019_accv, buhrmester_2019_IbPRIA, sachin_2018_SIRS, wu_2017_ICCCS}. Additionally, Buhrmester et al. \cite{buhrmester_2019_IbPRIA} investigated the effects of using RGB on a model trained with grayscale data, and found minimal impact on accuracy. Although they did not provide a detailed explanation, they speculated that essential visual cues such as edges and brightness are effectively learned and utilized by the model for object recognition tasks. For a short review of the impact of the color space on classification accuracy, see Velastegui et al. \cite{velastegui_2021_ICCSA}
\newline\newline
\noindent\textbf{Impact of color on human ability in face perception. }
The role of color in face perception by humans has been studied in Psychology. Early work by Bruce et al. \cite{bruce_1998_oup} and Kemp et al. \cite{kemp_1996_perception} largely dismissed the impact of color on face recognition. However, more recent work by Sinha et al. \cite{ sinha_2006_IEEE,sinha_2002_perecption}  demonstrated the influence of color on human ability in face detection and recognition. 
Brosseau et al. \cite{brosseau_2020_viscog}  reported that color-blind individuals performed significantly poorer on face recognition tasks, underscoring the importance of color.  Researchers have also explored the effects of both face and background color on the perception of facial expressions \cite{minami_2018_fip}. 
Bindemann et al. \cite{bindemann_2009_cogsci}  examined face detection performance in the absence of color and found that performance declines when color information is removed from faces, regardless of whether the surrounding scene context is rendered in color.
\newline\newline
\noindent\textbf{Impact of color on deep CNN face recognition.} 
Researchers have extensively examined the effects of blur, noise, occlusion, distortion, and other quality factors on the accuracy of deep CNN face recognition \cite{grm_2018_ietbiometrics, karahan_2016_biosig, majumdar_2021_iccv}. However, the influence of color has received less attention. To our knowledge, the work by Grm et al. \cite{grm_2018_ietbiometrics} is the only prior investigation on this specific topic, suggesting minimal accuracy difference on LFW test images for the models trained with and without color.
Our study differs from Grm et al. \cite{grm_2018_ietbiometrics} in five significant ways:
\begin{enumerate}
\item We employ deeper networks and utilize more advanced loss functions that are considered state-of-the-art for face recognition. Grm et al. \cite{grm_2018_ietbiometrics} evaluated several pre-ResNet architectures such as AlexNet, VGG-Face, GoogLeNet, and SqueezeNet, trained using simple categorical cross-entropy loss.  
\item We utilize larger datasets for training and multiple test datasets to evaluate differences, whereas \cite{grm_2018_ietbiometrics} utilized VGG-Face RGB images for training and tested only using LFW test images. 
\item We investigate the extent to which the network learns color-oriented features within the layer that analyzes the RGB input. By bottlenecking this layer and projecting a learned filter in the RGB space, we provide further insights into why models trained with grayscale images exhibit performance on par with those trained with color.
\item We analyze the extent to which the RGB values of the skin region vary across an individual's training images, and find that apparent skin tone is highly variable across in-the-wild images of a person.
\item We further analyze accuracy comparisons across four demographic groups (Caucasian Male/Female and African-American Male/Female) to assess whether RGB vs. grayscale images have varying impacts on individuals with different skin tones.
\end{enumerate}

\begin{figure*}[ht!]
  \begin{subfigure}[b]{1\linewidth}
    \centering
      \begin{subfigure}[b]{0.24\linewidth}
        \centering
          \includegraphics[width=1\linewidth]{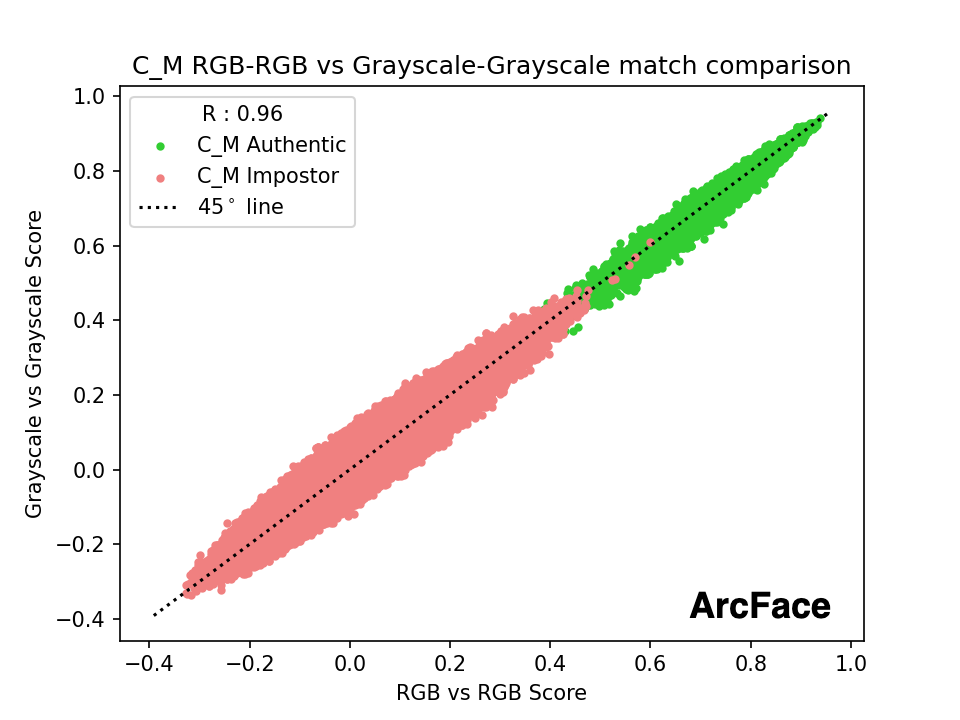}
      \end{subfigure}
      \hfill
      \begin{subfigure}[b]{0.24\linewidth}
        \centering
          \includegraphics[width=1\linewidth]{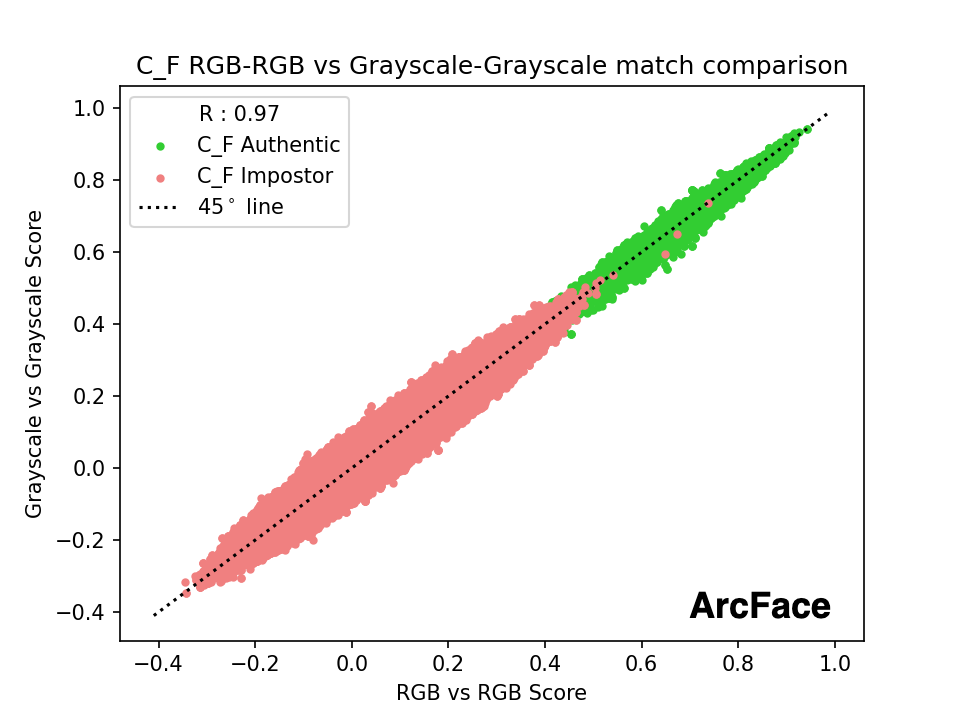}
      \end{subfigure}
      \hfill
      \begin{subfigure}[b]{0.24\linewidth}
        \centering
          \includegraphics[width=1\linewidth]{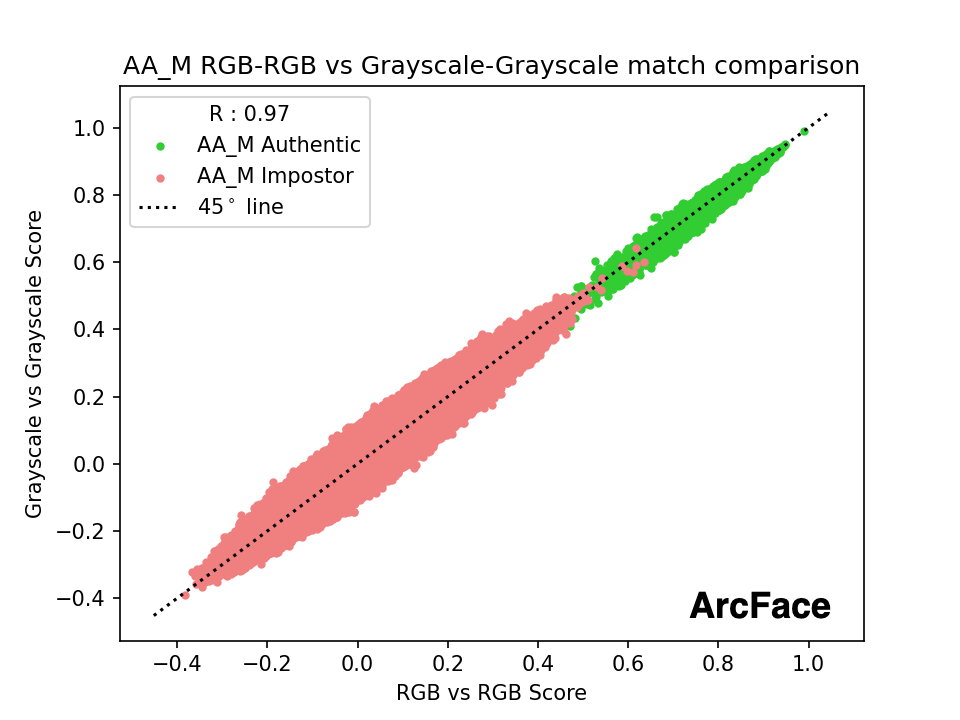}
      \end{subfigure}
      \hfill
      \begin{subfigure}[b]{0.24\linewidth}
        \centering
          \includegraphics[width=1\linewidth]{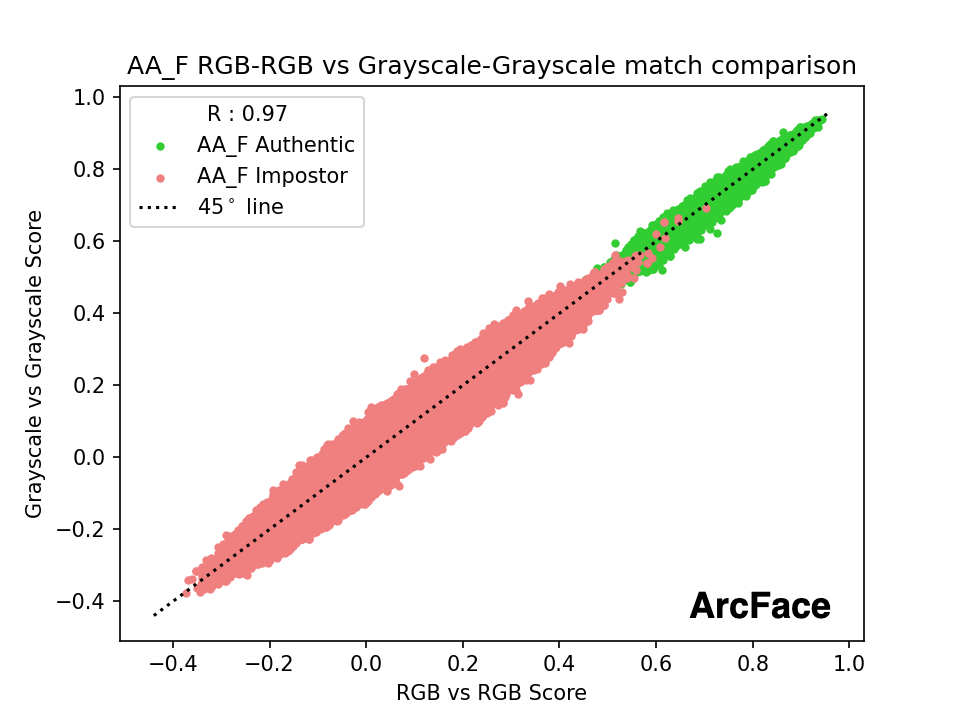}
      \end{subfigure}
  \end{subfigure}

  \begin{subfigure}[b]{1\linewidth}
    \centering
      \begin{subfigure}[b]{0.24\linewidth}
        \centering
          \includegraphics[width=1\linewidth]{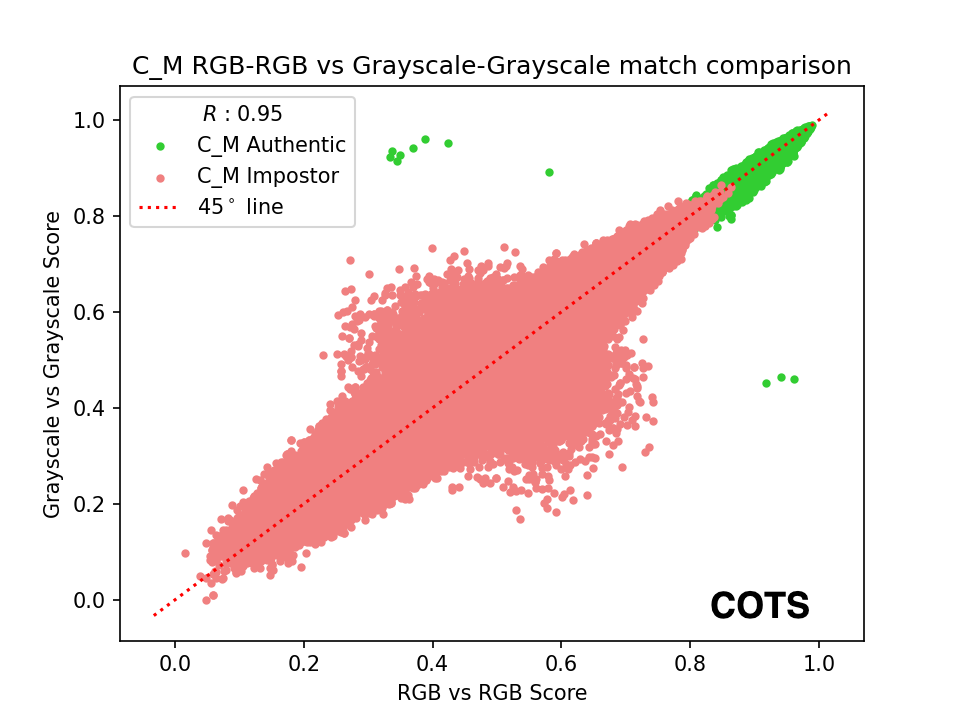}
          \caption{Caucasian Male}
      \end{subfigure}
      \hfill
      \begin{subfigure}[b]{0.24\linewidth}
        \centering
          \includegraphics[width=1\linewidth]{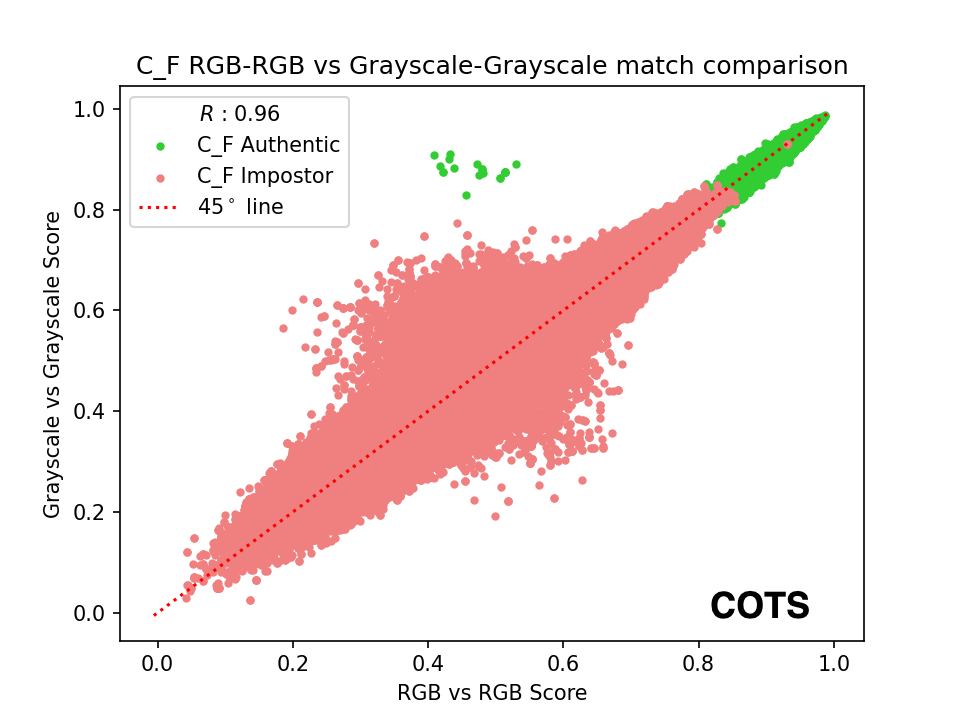}
          \caption{Caucasian Female}
      \end{subfigure}
      \hfill
      \begin{subfigure}[b]{0.24\linewidth}
        \centering
          \includegraphics[width=1\linewidth]{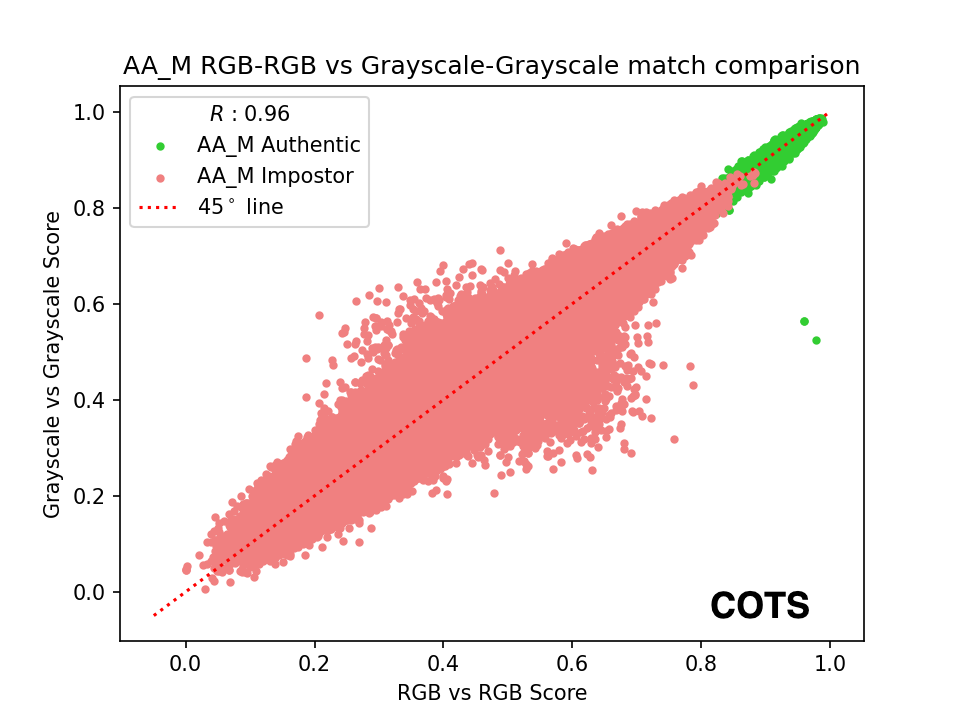}
          \caption{African-American Male}
      \end{subfigure}
      \hfill
      \begin{subfigure}[b]{0.24\linewidth}
        \centering
          \includegraphics[width=1\linewidth]{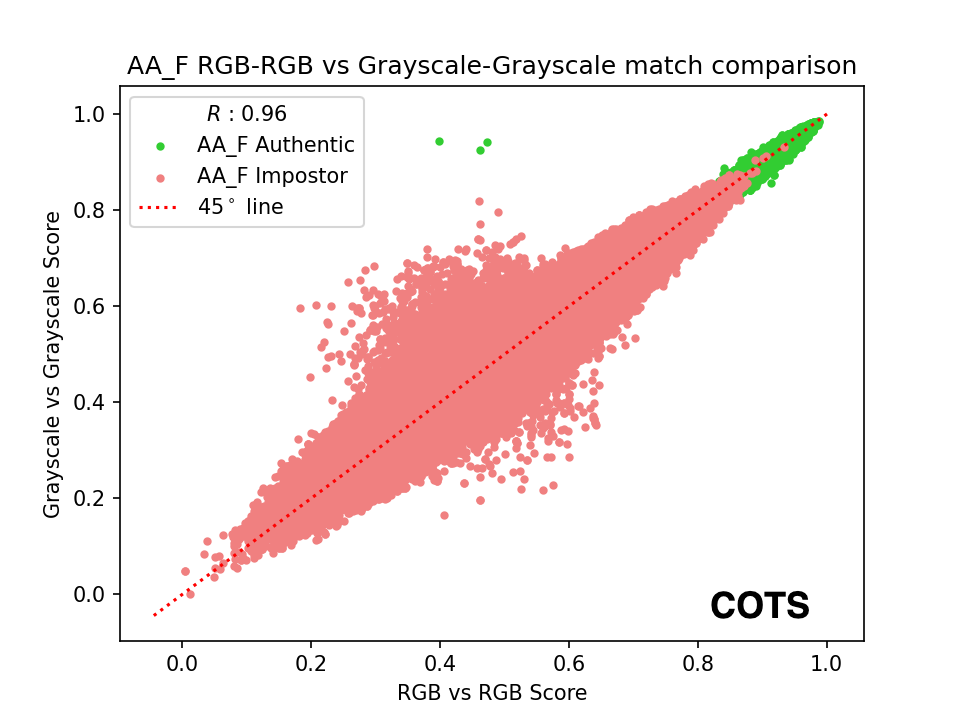}
          \caption{African-American Female}
      \end{subfigure}
  \end{subfigure}
  
  \caption{
   \textbf{Model Trained with RGB Images Exhibits Similar Performance When Applied To Grayscale Images From Diverse Demographics}. This suggests that using grayscale images do not disproportionately influence any specific demographic group. Each image pair presented in the plot has similarity score for original RGB version and grayscale version.  
For each demographic, throughout the range of similarity, the cloud of points trends on the 45-degree line.
If grayscale gave consistently lower similarity score, the cloud should trend below the 45-degree line.
{\it Top Row Figures - ResNet backbone, ArcFace loss, glint training set. Bottom Row Figures - COTS Matcher. Both tested on MORPH dataset}.
  }
  \label{fig:rgb-rgbvsgray-gray}
  \vspace{0.25em}
\end{figure*}
\begin{figure*}[h!]
    \centering
    \includegraphics[width=0.95\textwidth]{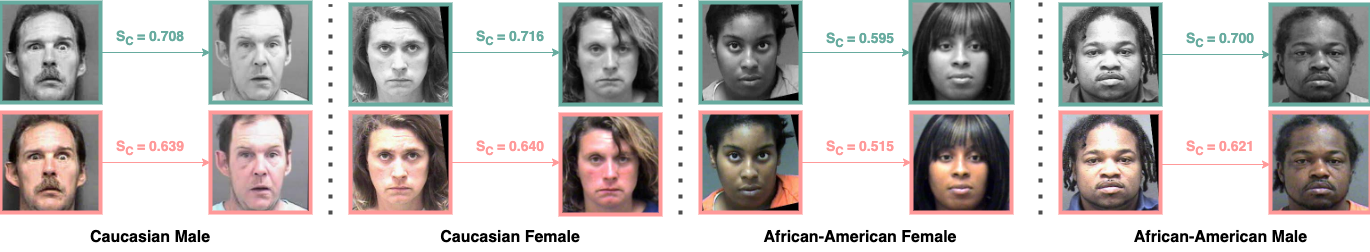}
    \caption{\textbf{Example image pairs from MORPH with grayscale similarity greater than RGB similarity.} Note that this result is from a matcher trained on RGB images. { \it Match Score ($S_c$) reported on ArcFace loss trained on Glint360k} \cite{Insightface} }
    \label{fig:rgb_vs_gray_sample}
    \vspace{-1em}
\end{figure*}

\section{Deep CNN face matchers do not ``see" color}

\subsection{Accuracy on Grayscale vs. RGB Test Sets}\label{pretrainedtest}

This section asks if popular color-trained face matchers achieve better accuracy for RGB vs grayscale test images. We use a combined-margin model based on ArcFace loss \cite{Deng_CVPR_2019}, trained on Glint360K (R100) \cite{an_iccv_2021} with weights from \cite{Insightface} and a Commercial-off-the-Shelf (COTS) matcher, not named due to license restrictions. 
For test images, we use the MORPH dataset \cite{morph_paper, morph_site}, consisting of 35,276 images of 8,835 Caucasian males, 10,941 images of 2,798 Caucasian females, 56,245 images of 8,839 African-American males, and 24,855 images of 5,928 African-American females.
The identities in MORPH should be fully disjoint from those in the web-scraped celebrity training sets such as Glint360K.

Figure \ref{fig:rgb-rgbvsgray-gray} presents a test of whether matchers have any consistent accuracy difference when tested on grayscale versus color versions of the same test images. 
The ArcFace matcher is trained on a known web-scraped, color image training set, and while the training set for the COTS matcher is not known, it is presumably at least as large as the union of current generally available training sets. 
The test sets are the four main demographic groups in  MORPH.  RGB-RGB image pairs are formed from the original MORPH color images. 
Corresponding gray-gray image pairs are formed by using OpenCV to create grayscale versions of the RGB images. 
{\it To ensure compatibility with the pre-trained model weights, grayscale images are loaded in a three-channel format with R=G=B.}
The RGB pairs and the grayscale pairs are processed by the same pretrained model that was trained on RGB images. 

Each pair of images corresponds to a point in Figure~\ref{fig:rgb-rgbvsgray-gray},
where cosine similarity for the RGB version of the pair is the horizontal axis and similarity for the grayscale version of the pair is the vertical axis.
If the similarity score is generally higher when matching RGB image pairs, the points would trend below the 45-degree line. If  similarity  is generally higher for grayscale, the points would trend above the 45-degree line.  The actual result is that the points cluster along the 45-degree line. This indicates that there is no consistent accuracy difference between matching grayscale versus color image pairs. As shown in Figure \ref{fig:rgb-rgbvsgray-gray}, the Pearson product-moment correlation coefficient (R) for the color and grayscale similarities is $\approx$ 0.97 for ArcFace and $\approx$ 0.96 for COTS. This is further quantitative evidence of the face similarity being essentially the same computed from color or grayscale. \\

\noindent{\bf Demographic differences consistent across color/grayscale.}
\noindent Figure~\ref{fig:rgb-rgbvsgray-gray} shows a consistent pattern of results across all four demographic groups for both matchers. Surprisingly, the use of color or grayscale does not appear to significantly enhance or degrade accuracy for any specific demographic. This suggests that 
factors beyond simply skin tone may underlie demographic accuracy differences in face recognition.

Points above the 45-degree line in Figure \ref{fig:rgb-rgbvsgray-gray} represent instances where the grayscale version of an image pair resulted in a higher similarity score than the color version.
These image pairs are intriguing given that the matchers were trained on color images.
Figure \ref{fig:rgb_vs_gray_sample} shows one example image pair per demographic group, highlighting the difference in similarity score between the color version and the grayscale version.

\subsection{Implementation Details for Training Face Matchers}\label{implementation}

This section summarizes the configuration for training the models in this work. 
We use two kinds of margin loss functions: one with a static margin value - ArcFace -  and the other with an adaptive margin value - AdaFace - to ensure that the findings are consistent for both paradigms. 
We use ResNet-18, ResNet-50, and ResNet-101 \cite{he_cvpr_2016} with modifications as proposed in \cite{Deng_CVPR_2019} as our baseline backbone. We train the models from scratch without using any weights from pretraining tasks.  For ArcFace, we use combined margin values of (1, 0, 0.4). We use cleaned WebFace4M dataset \cite{bhatta_wacvw_2024,zhu_CVPR_2021} as the training set. Images in WebFace4M are pre-aligned using RetinaFace \cite{deng_cvpr_2020retinaface}. The model is trained for 20 epochs using SGD as the optimizer \cite{paszke_nips_2019}, with momentum of 0.9, an initial learning rate of 0.1 and weight decay of 5e-4. We adopt polynomial decay as the learning rate scheduler during training from \cite{Insightface}. For AdaFace, we follow the original paper and use initial margin of 0.4. The model is trained for 26 epochs using SGD as the optimizer, with momentum of 0.9 and  initial learning rate of 0.1. The learning rate is reduced by a gamma factor of 0.1 at the 12th, 20th, and 24th epochs. Note that aside from adjusting the number of identities and the number of images for the ``color-cleaned" version (See $\S$ \ref{grayscaleimages} for definition), all other configuration values remain constant across all training instances.

\subsection{Grayscale vs. Color in the Training Data - Color as an additional information?}\label{grayscaleimages}

The importance of ``cleaner'' training data, in the sense of accurate identity labels, is widely acknowledged as  enabling training more accurate matchers. 
In this section, we show that (assumed) color training sets actually contain a fraction of grayscale images, and consider whether ``cleaning'' grayscale images out of an assumed color training set might help to explain the results in the previous section.

MS1MV2 \cite{Deng_CVPR_2019} is a cleaned version of MS-Celeb-1M \cite{guo_ECCV_2016}, and is widely used in training face matchers.
Glint360K \cite{an_iccv_2021} is newer than MS1MV2, and contains about 4x the number of identities and 3x the number of images.
WebFace4M/12M \cite{zhu_CVPR_2021} is newer still, and increasingly prominent, and its 4M subset has more than 2x the identities of MS1MV2 but fewer total images.
All three of these datasets contain web-scraped, in-the-wild images in RGB format.
We ran a test to detect images that have
R=G=B across all pixels, so that they are stored in RGB but effectively have grayscale content.
Results are summarized in Table~\ref{tab:grayscale_count}. Approximately 6-8\% of the images in each dataset are effectively grayscale. 
Even more importantly, 34\% (WebFace4M) to 60\% (MS1MV2) of the identities in each dataset have at least one grayscale image.
This is relevant because the goal of the deep CNN training is to get all images of each identity to classify as that identity.
Speculatively, having one or more grayscale images of an identity could lead the deep CNN to learn to ignore color in order to classify all images of an identity together.
\setlength\extrarowheight{2pt}
\begin{table}[ht]
\renewcommand{\arraystretch}{1.3}
\caption{\textbf{
Grayscale Image / Identities in Popular Training Sets.}
Similar to cleaning identity labels, cleaning is needed to avoid web-scraped RGB datasets having a large fraction of identities with one or more grayscale images.
}
\centering
\resizebox{0.975\columnwidth}{!}{%
\begin{tabular}{|c|cc|cc|}
\hline
\multicolumn{1}{|c|}{\multirow{2}{*}{Dataset}} & \multicolumn{2}{c|}{Original}      & \multicolumn{2}{c|}{Grayscale Subset}                       \\ \cline{2-5} 
\multicolumn{1}{|c|}{} &
  \multicolumn{1}{c|}{\begin{tabular}[c]{@{}l@{}}Total\\  Identities\end{tabular}} &
  \begin{tabular}[c]{@{}c@{}}Total\\ Images\end{tabular} &
  \multicolumn{1}{c|}{\begin{tabular}[c]{@{}l@{}}Identities w/ at least \\ one GrayScale image\end{tabular}} &
  \begin{tabular}[c]{@{}c@{}}Total \\ GrayScale Images\end{tabular} \\ \hline
MS1MV2                                         & \multicolumn{1}{l|}{$85.7$K} & $5.8$M  & \multicolumn{1}{l|}{$51.7$K \textcolor[HTML]{C41E3A}{( $\sim$$60$\% )}} & $444$K ( $\sim7.6$\% ) \\ \hline
Glint360k                                      & \multicolumn{1}{l|}{$360$K}  & $17.1$M & \multicolumn{1}{l|}{$154$K \textcolor[HTML]{C41E3A}{( $\sim$$43$\% )}}  & $919$K ( $\sim6$\% )   \\ \hline
WebFace4M                                      & \multicolumn{1}{l|}{$205$K}  & $4.2$M  & \multicolumn{1}{l|}{$70$K \textcolor[HTML]{C41E3A}{( $\sim$$34$\% )}}  & $246$K ( $\sim6$\% )   \\ \hline
\end{tabular}
}
\label{tab:grayscale_count}
\vspace{-0.2em}
\end{table}

To investigate how a fraction of the training images being effectively grayscale affects accuracy, we first create a ``color cleaned’’ version of the original WebFace4M by dropping images that have R=G=B. 
Excluding these effectively grayscale images leaves around 205K identities with 3.9M color images. This subset of WebFace4M is the ``\textbf{color cleaned}" version. 
We then train a network from scratch using each of (1) the color-cleaned (100\% color) subset of WebFace4M, and (2) the color-cleaned version of WebFace4M with all images converted to grayscale. We train an instance using ArcFace loss and one using AdaFace loss, for a total of four trained models. 
This allows a comparison of training on color versus grayscale of the same training images, for ArcFace and for AdaFace.
{\it Note that the grayscale image employed in this section of the experiment consists of three channels (R=G=B). This is to enable a direct comparison of gray versus color  without any modifications to the network architecture.} (Accuracy for the models trained with the original WebFace4M and its grayscale version can be found in supplementary material.) \\

\noindent{\bf Test for Statistical Significance.} Since we conduct multiple runs of the baseline experiments, we apply statistical tests to assess the significance of individual runs compared to these baselines. Our statistical analysis is conducted in two steps. Initially, we perform the {\bf Shapiro-Wilk test} to determine whether the underlying distribution of the accuracy values for various datasets for baseline runs follows a normal distribution. This is crucial because the assumption of normality is a prerequisite for conducting a standard t-test. The statistical test shows that there is not sufficient evidence to reject that the accuracy from multiple runs  follows a normal distribution. 
We then proceed to perform a one-sample t-test to measure the statistical significance of the observed accuracy in grayscale training, comparing it to the baseline RGB accuracies. {\bf A one-sample t-test} is a statistical method used to determine whether the mean of a single sample significantly differs from a known or hypothesized population mean. This involves calculating a t-statistic from the sample data, which is computed by dividing the difference between the sample mean and the population mean by the standard error of the mean (the standard deviation divided by the square root of the sample size). The p-value is then derived by comparing this t-statistic to the t-distribution corresponding to the degrees of freedom (sample size minus one), which quantifies the probability of observing a value as extreme or more extreme by chance, assuming the null hypothesis is true. We use the commonly accepted threshold for a p-value in statistical tests, which is 0.05. This threshold value signifies that if the p-value is less than 0.05, there is less than a 5\% chance that the observed results happened by chance under the null hypothesis (i.e., the observation comes from the same distribution as the previous data), which suggests a significant difference is observed. \\

\noindent{\bf Evaluation on Standard Benchmarks.}
Table~\ref{tab:result1} summarizes these accuracy comparisons for training on grayscale versus color. The baseline model, trained with color, was run independently five times, reporting mean accuracy and standard deviation. For the model trained with three-channel grayscale images, accuracy along with statistical significance estimates comparing it to baseline runs is reported. Accuracy is listed for each of the validation datasets: LFW \cite{lfw}, CFP-FP \cite{cfpfp}, AGEDB-30 \cite{AGEDB30}, CALFW \cite{calfw}, and CPLFW \cite{cplfw}. Note that all of these validation datasets contain primarily color images.

A consistent pattern emerges From Table~\ref{tab:result1}. For shallower models, like ResNet-18, the model trained with grayscale consistently exhibits  statistically significantly lower accuracy than the model trained with color images. However, the differences between color and grayscale models diminish as deeper models are used. In the case of slightly deeper networks, such as ResNet-50, the model trained with grayscale images begins to achieve comparable performance to training with color. The average accuracy for the grayscale model trained with ArcFace loss falls within the statistical bounds of the color model, and for the model trained with AdaFace, the accuracy gap is significantly reduced compared to shallower counterparts. Finally, for deeper ResNet-101, there are no statistically significant accuracy differences between the model trained with grayscale or color for standard face recognition benchmarks. {\it This is especially intriguing because the model trained with grayscale has not encountered color during its training, yet it performs equally well on color test images as the model trained with color images.}

We hypothesize that this behavior primarily arises from the varying capacities of the models to represent features. Shallower networks inherently possess a lower capacity to learn complex feature representations. As grayscale images contain less information, these shallower networks appear to rely more heavily on low-level color features, potentially limiting their ability to capture intricate patterns from the grayscale image. \\

\setlength\extrarowheight{3pt}
\begin{table}[!ht]
\renewcommand{\arraystretch}{1.3}
\caption{ Performance comparison of {\bf three-channel grayscale (R=G=B) and RGB training} using {\bf RGB test images} for different model depths.
Training on three-channel grayscale images (R=G=B) and testing on RGB images yields comparable performance to RGB training and testing for deeper models, but shows differences for shallower models.
Key: \textcolor[HTML]{C41E3A}{$p \textcolor[HTML]{C41E3A}{\textless 0.05} $ ; lower accuracy + significant difference}, \textcolor[HTML]{006400}{$p \textgreater 0.05$ ; no significant difference or higher accuracy}
}
\centering
\resizebox{\columnwidth}{!}{
\begin{tabular}{c|c|ccc||ccc}
\hline
          &  Loss $\rightarrow$          &     & ArcFace &         &     & AdaFace &         \\ \hline
Backbone  & \begin{tabular}[c]{@{}c@{}}\diagbox{Test $\downarrow$}{Train $\rightarrow$ }\end{tabular} & RGB(3) & Gray(3)     & p-value & RGB(3) & Gray(3)  & p-value \\ \hline
          & LFW        & 99.60 $\pm$ 0.08  & 99.56        & \textcolor[HTML]{006400}{0.32}        & 99.57 $\pm$ 0.03    &  99.47   & \textcolor[HTML]{C41E3A}{\textless 0.05}        \\
          & CFP-FP     & 97.71 $\pm$ 0.18  &  97.28       & \textcolor[HTML]{C41E3A}{\textless 0.05}   & 97.27 $\pm$ 0.11  &  96.51       & \textcolor[HTML]{C41E3A}{\textless 0.05}    \\
ResNet-18 & AGEDB-30   & 96.63 $\pm$ 0.19  & 96.48     & \textcolor[HTML]{006400}{0.15}      & 96.24 $\pm$ 0.18    & 95.95       & \textcolor[HTML]{C41E3A}{\textless 0.05}         \\
          & CALFW      & 95.58 $\pm$ 0.08 &  95.43       & \textcolor[HTML]{C41E3A}{\textless 0.05}         & 95.35 $\pm$ 0.20    & 95.23        & \textcolor[HTML]{006400}{0.25}         \\
          & CPLFW      & 92.39 $\pm$ 0.17   & 91.88      & \textcolor[HTML]{C41E3A}{\textless 0.05}         & 91.95 $\pm$ 0.07    & 91.40       & \textcolor[HTML]{C41E3A}{\textless 0.05}         \\ \cdashline{2-8}
          & Average    & 96.38 $\pm$ 0.07 &  96.13       & \textcolor[HTML]{C41E3A}{\textless 0.05}         & 96.08 $\pm$ 0.09    &  95.71       & \textcolor[HTML]{C41E3A}{\textless 0.05}         \\ \hline\hline
          
          & LFW        & 99.77 $\pm$ 0.03   &  99.80      & \textcolor[HTML]{006400}{0.09}        & 99.77 $\pm$ 0.05    & 99.78        & \textcolor[HTML]{006400}{0.67}        \\
          & CFP-FP     & 99.03 $\pm$ 0.06  &  98.83      & \textcolor[HTML]{C41E3A}{\textless 0.05}         & 98.76 $\pm$ 0.04    & 98.73       & 0.17         \\
ResNet-50 & AGEDB-30   & 97.56 $\pm$ 0.14   & 97.63      & \textcolor[HTML]{006400}{0.32}        & 97.48 $\pm$ 0.05    & 97.29        & \textcolor[HTML]{C41E3A}{\textless 0.05}        \\
          & CALFW      & 95.98 $\pm$ 0.08  & 95.98       & \textcolor[HTML]{006400}{1.0}       & 95.96 $\pm$ 0.06    & 95.85        & \textcolor[HTML]{C41E3A}{\textless 0.05}          \\
          & CPLFW      & 94.07 $\pm$ 0.10  &  93.68       & \textcolor[HTML]{C41E3A}{\textless 0.05}          & 93.87 $\pm$ 0.1    & 93.75        & 0.06    \\\cdashline{2-8}
          & Average    & 97.28 $\pm$ 0.08  &  97.19       & \textcolor[HTML]{006400}{0.07}       & 97.18 $\pm$ 0.05    & 97.10        &  \textcolor[HTML]{C41E3A}{\textless 0.05}        \\\hline\hline

          & LFW        & 99.78 $\pm$ 0.03    & 99.80       & \textcolor[HTML]{006400}{0.21}        & 99.80 $\pm$ 0.03    &  99.75       & \textcolor[HTML]{C41E3A}{\textless 0.05}         \\
          & CFP-FP     & 99.18 $\pm$ 0.05    &  99.17       & \textcolor[HTML]{006400}{0.67}        & 99.07 $\pm$ 0.09    & 98.95        & \textcolor[HTML]{006400}{0.05}        \\
ResNet-101& AGEDB-30   & 97.99 $\pm$ 0.10    &  97.87       & \textcolor[HTML]{006400}{0.06}        & 97.74 $\pm$ 0.04   & 97.83       &  \textcolor[HTML]{C41E3A}{\textless 0.05}        \\
          & CALFW      & 96.06 $\pm$ 0.07    &  96.02       & \textcolor[HTML]{006400}{0.27}        & 96.02 $\pm$ 0.12   & 95.97        & \textcolor[HTML]{006400}{0.40}      \\
          & CPLFW      & 94.42 $\pm$ 0.09    &  94.52      & \textcolor[HTML]{006400}{0.07}        & 94.34 $\pm$ 0.09  & 94.12        & \textcolor[HTML]{C41E3A}{\textless 0.05}         \\\cdashline{2-8}
          & Average    & 97.45 $\pm$ 0.11    &  97.48       & \textcolor[HTML]{006400}{0.57}        & 97.40 $\pm$ 0.06   & 97.33        & \textcolor[HTML]{006400}{0.06}      \\\hline
\end{tabular}
}

\label{tab:result1}
\end{table}

\noindent{\bf Evaluation on IJB-B and IJB-C.}
IJB-B \cite{white_cvprw_2017} and IJB-C \cite{maze_ijcb_2018} are datasets recognized for increased difficulty compared to the standard benchmarks. 
Evaluation on IJB-B and IJB-C is particularly important  due to Sinha et al.'s findings \cite{sinha_2002_perecption}, which highlight the increased importance of color cues in situations where shape cues are degraded.

The TAR@FAR=0.01\% results are summarized in Table \ref{tab:ijb}.
Note that the test images used in our experiments are color images. 
The pattern observed on the standard benchmark datasets is consistently replicated in the IJB suite.
As model capacity increases, the accuracy differences between models trained with grayscale and color diminish. Specifically, for ArcFace with the deeper ResNet101, the performance of the grayscale model falls within the statistical bounds of the color model. However, for AdaFace, although the accuracy on grayscale is lower, the difference between the color and grayscale models is much reduced compared to shallower models.

In contrast to how humans utilize colors for recognition in low-quality images \cite{sinha_2002_perecption}, our results demonstrate that deeper models, with their ability to represent intricate features even with low-level information, perform comparably well when {\it trained on grayscale images and tested on color images}, even in a low-quality test setting.

\setlength\extrarowheight{3pt}
\begin{table}[!ht]
\renewcommand{\arraystretch}{1.3}
\caption{ Training on {\bf three-channeled grayscale (R=G=B) or color images} yields consistent results as standard benchmark test sets for models of varying depths when tested on {\bf color images}, even in low-quality test image settings. As the model depth increases, the accuracy gap between RGB and grayscale training reduces. 
Key: \textcolor[HTML]{C41E3A}{$p \textcolor[HTML]{C41E3A}{\textcolor[HTML]{C41E3A}{\textless 0.05} } $ ; lower accuracy + significant difference}, \textcolor[HTML]{006400}{$p \textgreater 0.05$ ; no significant difference}
}
\centering
\resizebox{\columnwidth}{!}{
\begin{tabular}{c|c|ccc||ccc}
\hline
          &  Loss $\rightarrow$          &     & ArcFace &         &     & AdaFace &         \\ \hline
Backbone  & \begin{tabular}[c]{@{}c@{}}\diagbox{Test $\downarrow$}{Train $\rightarrow$ }\end{tabular} & RGB(3) & Gray(3)     & p-value & RGB(3) & Gray(3)  & p-value \\ \hline
ResNet-18 & IJB-B    & 93.12 $\pm$ 0.07   & 92.52        &  \textcolor[HTML]{C41E3A}{\textless 0.05}        & 93.06 $\pm$ 0.09    & 92.35       &  \textcolor[HTML]{C41E3A}{\textless 0.05}         \\
          & IJB-C    & 95.16 $\pm$ 0.02    & 94.57        & \textcolor[HTML]{C41E3A}{\textless 0.05}          & 94.95 $\pm$ 0.05    & 94.42       &   \textcolor[HTML]{C41E3A}{\textless 0.05}        \\ \hline
ResNet-50 & IJB-B    &95.16 $\pm$ 0.05    &  95.05       &  \textcolor[HTML]{C41E3A}{\textless 0.05}         & 95.26 $\pm$ 0.13    & 94.96        & \textcolor[HTML]{C41E3A}{\textless 0.05}          \\ 
          & IJB-C    &96.88 $\pm$ 0.03    &  96.78       &  \textcolor[HTML]{C41E3A}{\textless 0.05}         & 96.77 $\pm$ 0.08     & 96.64        & \textcolor[HTML]{C41E3A}{\textless 0.05}          \\ \hline
ResNet-101& IJB-B     & 95.71 $\pm$ 0.08     & 95.68        & \textcolor[HTML]{006400}{0.44}        & 95.72 $\pm$ 0.11   & 95.37        & \textcolor[HTML]{C41E3A}{\textless 0.05}         \\ 
          & IJB-C     & 97.28 $\pm$ 0.06    &  97.24       & \textcolor[HTML]{006400}{0.21}        & 97.13 $\pm$ 0.09    &  96.97       & \textcolor[HTML]{C41E3A}{\textless 0.05}         \\ \hline
\end{tabular}
}
\vspace{-0.5em}
\label{tab:ijb}
\end{table}

\section{Insights into Achromatopsia within Face Recognition Networks}
The previous section presents a quantitative analysis of model performance using both RGB and three-channel grayscale (R=G=B) training sets. This section shifts to a more qualitative exploration of the role of color in face recognition networks.

\subsection{Is HSV Color Space Better Than RGB?}\label{colorspace}
RGB is the universal format for color images in face recognition pipelines.
However, our results above may motivate the question of whether a different color space could enable the network to learn more from color images.
RGB can be viewed as having the disadvantage of not explicitly separating chromaticity and luminosity. 
HSV (Hue, Saturation, Value) can be viewed as having a possible advantage in that luma information is isolated in one plane (value), and chroma (color) in the other two (hue and saturation). 
To investigate whether separating chroma and luma can enable the network to learn more from color, we ran a parallel set of experiments with color images converted to HSV for training and testing. Since RGB and HSV both consist of three channels, no change is needed in the CNN architecture. 

We trained and tested ArcFace and AdaFace using an HSV version of the ``color-cleaned" subset of WebFace4M. Results in Table \ref{tab:hsv} show that HSV does not consistently result in better or worse accuracy than RGB, except for the shallower ResNet-18 model, where the model trained and tested on HSV is consistently lower. %
We hypothesize that shallower models like ResNet-18 are unable to adeptly capture the representation shift from RGB to HSV. HSV encodes color information differently, adding additional relationships between Hue, Saturation, and Value that the models need to learn. 
A more detailed analysis of which HSV planes the HSV-trained model primarily relies on is presented in the next section.

\setlength\extrarowheight{3pt}
\begin{table}[!ht]
\renewcommand{\arraystretch}{1.3}
\caption{Separating the ``chroma" and ``luma" information does not neccesarily change the pattern of results when compared to training in the RGB color space or using grayscale images. 
Key: \textcolor[HTML]{C41E3A}{$p \textcolor[HTML]{C41E3A}{\textless 0.05} $ ; lower accuracy + significant difference}, \textcolor[HTML]{006400}{$p \textgreater 0.05$ ; no significant difference or higher accuracy}
}
\centering
\resizebox{\columnwidth}{!}{
\begin{tabular}{c|c|ccc||ccc}
\hline
          &  Loss $\rightarrow$          &     & ArcFace &         &     & AdaFace &         \\ \hline
Backbone  & \begin{tabular}[c]{@{}c@{}}\diagbox{Test $\downarrow$}{Train $\rightarrow$ }\end{tabular} & RGB(3) & HSV(3)     & p-value & RGB(3) & HSV(3)  & p-value \\ \hline
          & LFW        & 99.60 $\pm$ 0.08  & 99.66        & \textcolor[HTML]{006400}{0.17}       & 99.57 $\pm$ 0.03    &  99.58       & \textcolor[HTML]{006400}{0.49}        \\
          & CFP-FP     & 97.71 $\pm$ 0.18  &  97.31       & \textcolor[HTML]{C41E3A} {\textless 0.05}        & 97.27 $\pm$ 0.11    &  97.05       &  \textcolor[HTML]{C41E3A} {\textless 0.05}       \\
ResNet-18 & AGEDB-30   & 96.63 $\pm$ 0.19  & 96.35        & \textcolor[HTML]{C41E3A} {\textless 0.05}        & 96.24 $\pm$ 0.18    & 96.03       & \textcolor[HTML]{006400}{0.07}         \\
          & CALFW      & 95.58 $\pm$ 0.08 &  95.45       & \textcolor[HTML]{C41E3A} {\textless 0.05}        & 95.35 $\pm$ 0.20    & 95.34        & \textcolor[HTML]{006400}{0.91}         \\
          & CPLFW      & 92.39 $\pm$ 0.17   & 91.88        & \textcolor[HTML]{C41E3A} {\textless 0.05}        & 91.95 $\pm$ 0.07    & 91.78       & \textcolor[HTML]{C41E3A} {\textless 0.05}         \\ \cdashline{2-8}
          & Average    & 96.38 $\pm$ 0.07 &  96.13       & \textcolor[HTML]{C41E3A} {\textless 0.05}        & 96.08 $\pm$ 0.09    &  95.96       & \textcolor[HTML]{C41E3A} {\textless 0.05}         \\ \hline\hline

          & LFW        & 99.77 $\pm$ 0.03   &  99.82       & \textcolor[HTML]{C41E3A} {\textless 0.05}        & 99.77 $\pm$ 0.05    & 99.78        & \textcolor[HTML]{006400}{0.44}         \\
          & CFP-FP     & 99.03 $\pm$ 0.06  &  98.97      & \textcolor[HTML]{006400}{0.09}        & 98.76 $\pm$ 0.04    & 98.60       & \textcolor[HTML]{C41E3A} {\textless 0.05}         \\
ResNet-50 & AGEDB-30   & 97.56 $\pm$ 0.14   & 97.70        &  \textcolor[HTML]{006400}{0.09}       & 97.48 $\pm$ 0.05    & 97.30        & \textcolor[HTML]{C41E3A} {\textless 0.05}        \\
          & CALFW      & 95.98 $\pm$ 0.08  & 96.01        &  \textcolor[HTML]{006400}{0.45}       & 95.96 $\pm$ 0.06    & 96.02        & \textcolor[HTML]{006400}{0.09}         \\
          & CPLFW      & 94.07 $\pm$ 0.10  &  94.20       & \textcolor[HTML]{006400}{0.05}        & 93.87 $\pm$ 0.1    & 94.02        & \textcolor[HTML]{C41E3A} {\textless 0.05}         \\\cdashline{2-8}
          & Average    & 97.28 $\pm$ 0.08  &  97.34       &  \textcolor[HTML]{006400}{0.17}       & 97.18 $\pm$ 0.05    & 97.14        &  \textcolor[HTML]{006400}{0.15}       \\\hline\hline

          & LFW        & 99.78 $\pm$ 0.03    & 99.83        & \textcolor[HTML]{C41E3A} {\textless 0.05}        & 99.80 $\pm$ 0.03    &  99.80       & \textcolor[HTML]{006400}{1.0}        \\
          & CFP-FP     & 99.18 $\pm$ 0.05    &  99.19       & \textcolor[HTML]{006400}{0.67}        & 99.07 $\pm$ 0.09    & 99.04        & \textcolor[HTML]{006400}{0.50}        \\
ResNet-101& AGEDB-30   & 97.99 $\pm$ 0.10    &  98.08       & \textcolor[HTML]{006400}{0.12}        & 97.74 $\pm$ 0.04   & 97.98        & \textcolor[HTML]{006400} {\textless 0.05}        \\
          & CALFW      & 96.06 $\pm$ 0.07    &  95.90       & \textcolor[HTML]{C41E3A} {\textless 0.05}        & 96.02 $\pm$ 0.12   & 95.92        & \textcolor[HTML]{006400}{0.14}        \\
          & CPLFW      & 94.42 $\pm$ 0.09    &  94.38      & \textcolor[HTML]{006400}{0.37}        & 94.34 $\pm$ 0.09  & 94.33        & \textcolor[HTML]{006400}{0.81}        \\\cdashline{2-8}
          & Average    & 97.45 $\pm$ 0.11    &  97.47       & \textcolor[HTML]{006400}{0.70}        & 97.40 $\pm$ 0.06   & 97.42        & \textcolor[HTML]{006400}{0.50}       \\\hline
\end{tabular}
}
\label{tab:hsv}
\end{table}

\begin{figure*}[ht!]
 \centering
  \begin{subfigure}[b]{\linewidth}
      \begin{subfigure}[b]{0.325\linewidth}
        \centering
          \includegraphics[width=1\linewidth]{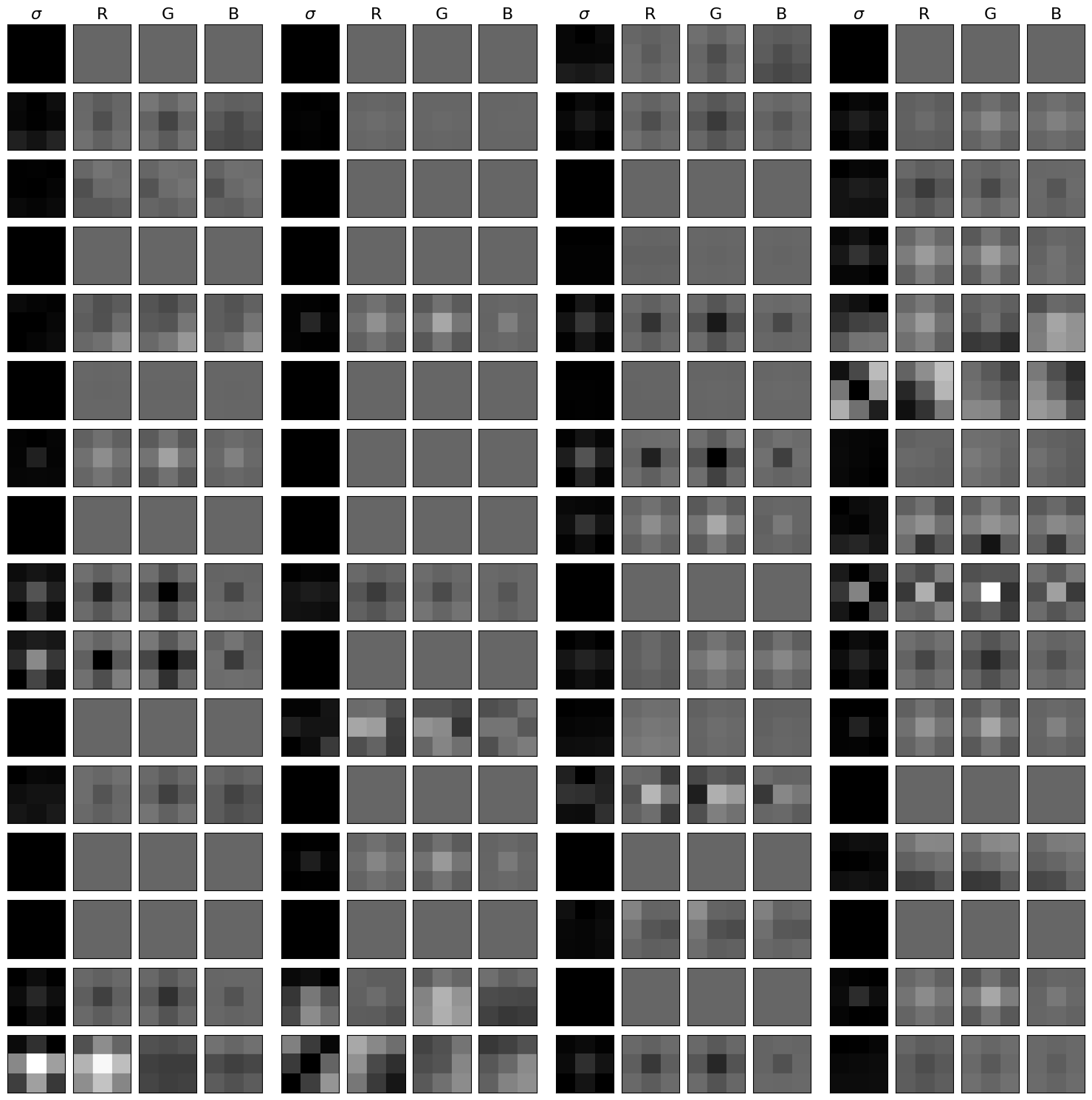}
          \caption{\scriptsize{``color-cleaned" WebFace4M (on RGB)}}\label{wf4mrgb}
      \end{subfigure}
      \hfill
      \begin{subfigure}[b]{0.325\linewidth}
        \centering
          \includegraphics[width=1\linewidth]{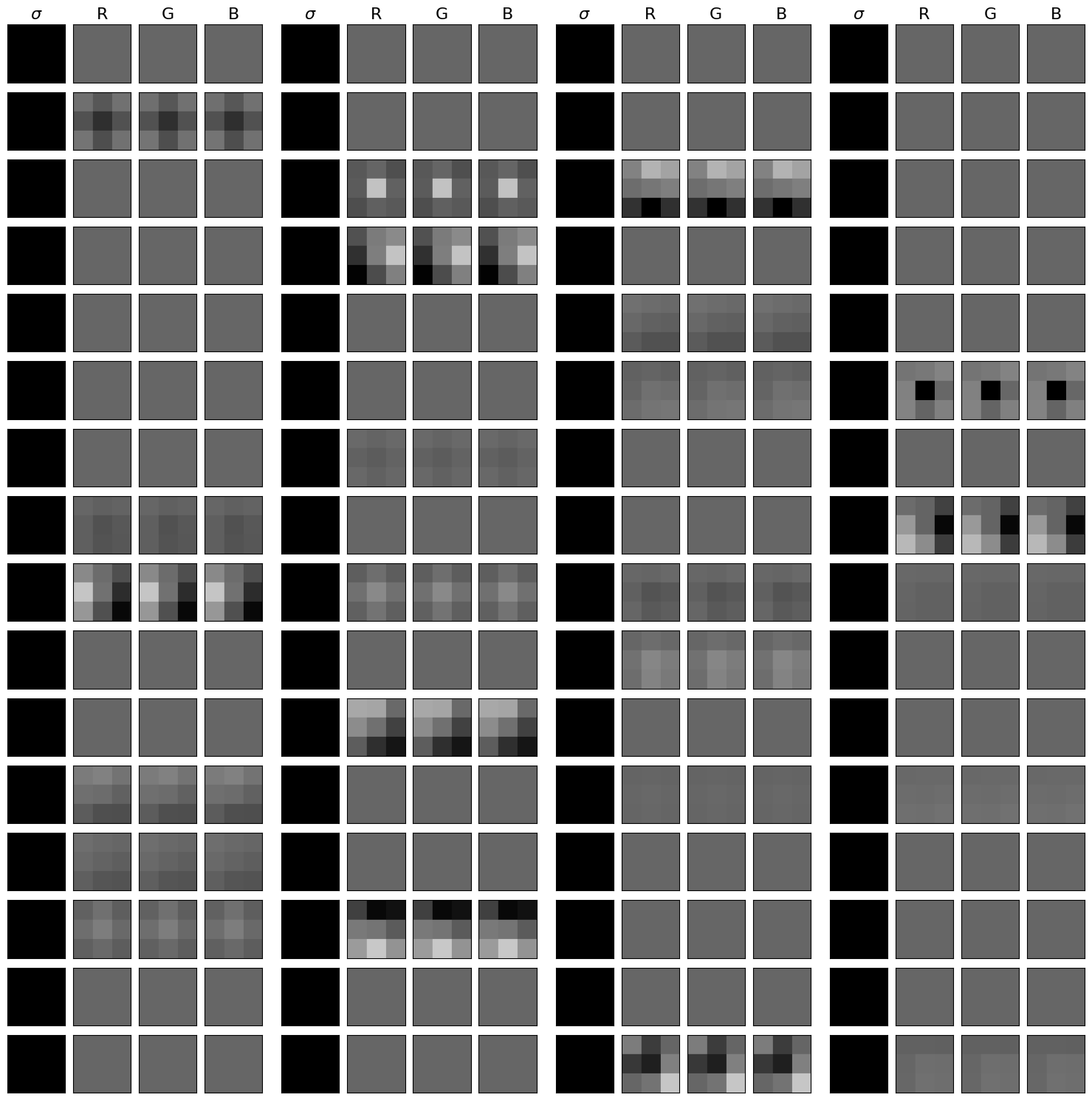}
          \caption{\scriptsize{``color-cleaned" WebFace4M (on Gray 3-channel)}}\label{wf4mgray}
      \end{subfigure}
      \hfill
      \begin{subfigure}[b]{0.325\linewidth}
        \centering
          \includegraphics[width=1\linewidth]{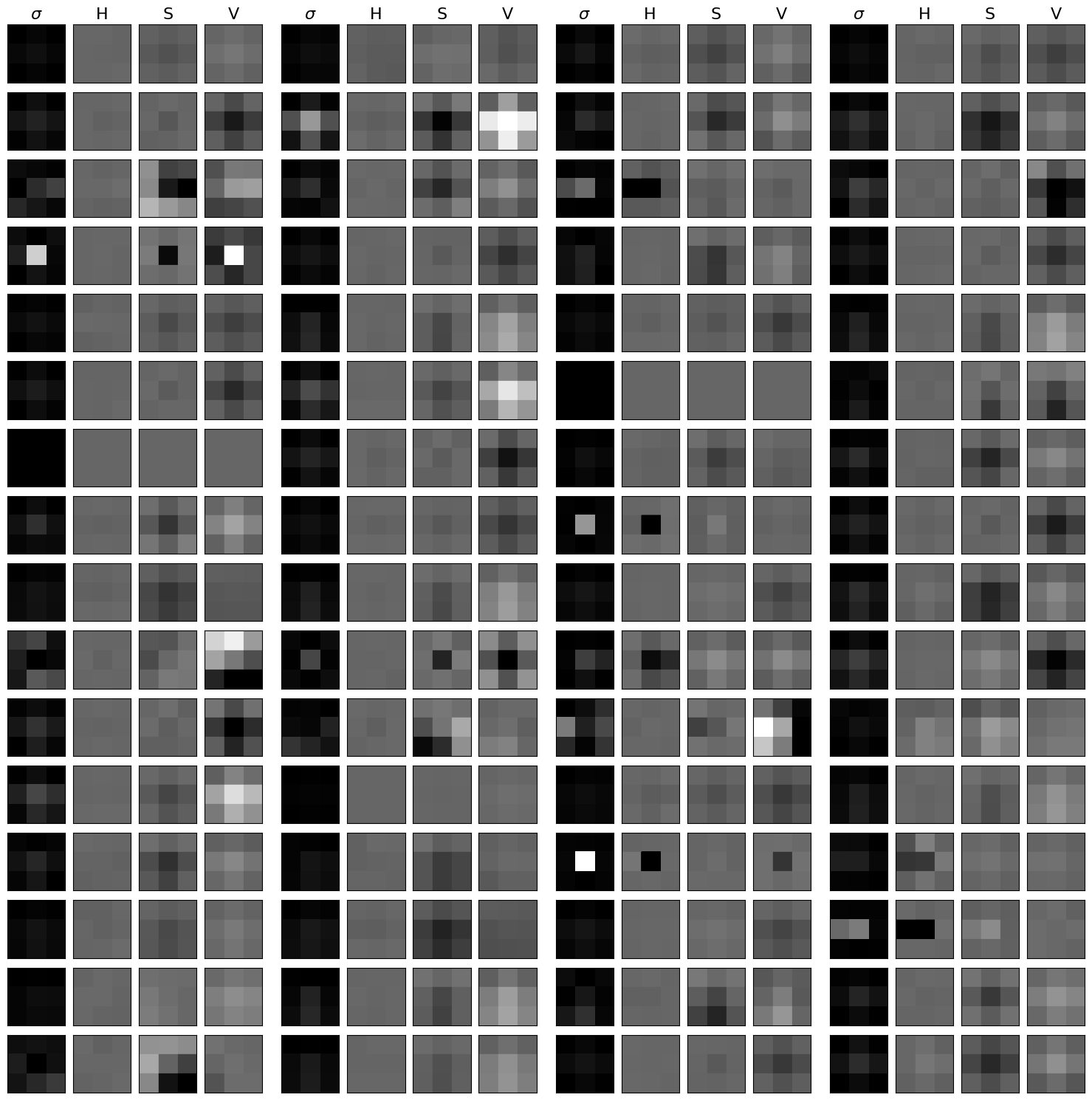}
          \caption{\scriptsize{``color-cleaned" WebFace4M (on HSV)}}\label{wf4mhsv}
      \end{subfigure}
      
  \end{subfigure}
  \caption{\textbf{Visualization of 64 Convolution Filter Weight Values of the First Convolution Block in row-major order.} Approximately one-third of the convolution blocks dedicated to RGB data reached nearly zero values, while the remaining blocks exhibited a strikingly similar pattern across the RGB planes. However, only a few convolution blocks displayed significantly different values for the RGB planes. In contrast, when considering the HSV plane, it appears that the most active block primarily derives its information from the V plane, indicating that the network learns to extract more valuable data from this particular plane compared to the others. In three-channel grayscale training, all three planes contain the same information, which results in same values across the planes for all filters. \it{ Backbone: ResNet-50, Loss: ArcFace } }
  \label{fig:weight_vis}
  \vspace{-1em}
\end{figure*}

\subsection{Color in First Convolutional Layer Weights}\label{deeperlook}

In this section, we analyze what the network learns about using color by visualizing the pattern of weights learned in the first convolutional layer for ArcFace. 
We chose the ResNet-50 model for this analysis because it is not as shallow as ResNet-18, which may not adequately capture intricate features from low-level data, nor as deep as ResNet-101, which is highly capable of handling various complexities. (Visualization and discussion for the ResNet-50 and ResNet-101 can be found in the supplementary material. )   We selected the model trained with ArcFace loss because it lacks quality adaptive components like AdaFace, making it more suitable for weight visualization and easier to interpret. 

For the ResNet backbone, the color image is input to the first convolutional layer, which learns 64 different $3 \times 3 \times 3$ convolutions; $3 \times 3 \times 3$ because it is a $3 \times 3$ kernel applied to each of R, G and B.
Each of the 64 filters can learn to extract a different feature image from the image.
So, for each of the 64 convolutions learned in the first layer, there is a $3 \times 3$ pattern of weights for each of the R, G and B color planes.
After the first layer, the learned weights are no longer directly tied to the color planes.

We visualize the $3 \times 3 \times 3$ learned weights for a given one of the 64 convolutions through a set of four $3 \times 3$ grayscale grids. The first grid represents the standard deviation of the values across the R, G, and B weights at each pixel position.  
An all-black $3 \times 3$ grid in the first column shows that the weights are the same across R, G and B; in effect, the learned filter is extracting grayscale information.
White in a $3 \times 3$ grid in the first column represents the maximum standard deviation across the R, G, and B weights among all 64 sets of $3 \times 3 \times 3$ weights. The second, third and fourth grids represent the weights for the R, G and B planes, respectively.  The weights are linearly scaled for better visualization. In these three columns, negative weights are depicted as black, zero weights as gray, and positive weights as white. This scheme allows us to visualize characteristics of the learned filter weights.

Figure~\ref{fig:weight_vis} contains the visualization of the weights for training ArcFace on the RGB color-cleaned subset of WebFace4M on the left side, and the weights for training ArcFace on the HSV version of the same images on the right side.
Each set of 64 convolution weights is shown as 4 columns of 16.
Consider the visualization of the convolution weights for ArcFace trained on RGB, as shown in Figure \ref{wf4mrgb}.
The visualization of convolution weights in the upper left corner shows that the standard deviation of the weights is zero (black in the first $3 \times 3$ grid) and that the weights in the $3 \times 3$ convolution are zero for each of the R, G and B color planes (grey in the other $3 \times 3$ grids).
This is an example of a filter that converged to a convolution that just produces the value zero.
About one third of the 64 learned convolutions converged to a similar result, as observed in \cite{bhatta_craft_arxiv_2023}.
Most of the remaining convolutions have a weight pattern that is  similar across R, G and B.
A similar pattern of weights across R, G and B would produce a result  similar to applying the average weights on a grayscale version of the image.
Only a few of the convolutions have a pattern of weights that appears substantially different across R, G and B.
For example, the convolution in the lower left in Figure \ref{wf4mrgb} appears to be a spot or line detector using primarily the R plane. In the model trained with three-channel grayscale as shown in Figure \ref{wf4mgray}, all the planes visually appear to have learned the same pattern of weights. This occurs because training in three-channel grayscale provides identical information across all planes. Additionally, upon examining the standard deviation column for all filters, it is observed that 
all three planes possess near identical weights.

Now, contrast the visualization of the weights for the RGB-trained ArcFace with those for the HSV-trained ArcFace, shown in  Figure~\ref{wf4mhsv}.
In the visualization of these 64 convolutions, there is much more variation in the pattern of weights across H, S and V.  The weights are generally near zero across the $3 \times 3$ grid for H.  There is generally more variation in the pattern of weights for S.  And by far the greatest overall variation is in V.
Thus, the HSV-learned weights suggest that the network learns to extract more information from the V plane, which is effectively just the grayscale, than from the other two planes.

Visualization of the convolution weights learned for either  RGB  or  HSV  indicates that the network extracts relatively little color-oriented information to make available to later layers. 
For the RGB convolutions, only a few learn substantially different weights for R, G and B.
For the HSV convolutions, most of the variation in weights is focused on the V plane, and relatively little is extracted from H.
The next section presents an analysis of the color of the skin region for selected identities in the training set that suggests why this might be the case.

\subsection{Color Variation Within an Identity's Images. }\label{vartrainingdata}

The network should learn to extract color-oriented features from the the training images to the degree that such features aid in classifying images of an identity into the same ``bin’’.  
Based on this perspective, we analyze images from selected identities in the training set to illustrate how useful, or not, the color information in the training images could be.

Now consider a probability distribution for $n$ classes, where each class has the same number of data points and equal probabilities. This can be represented as follows:
\begin{equation}
    P(X = x_i) = \frac{1}{n}, \quad i = 1, 2, \ldots, n
\end{equation}

If  color is an essential feature to distinguish between identities, the nearest neighbor image of a given identity should be another image of the same identity in the RGB space. If color was a purely random clue, then with $n$ identities, the probability of point’s nearest neighbor would be from the same identity about $1/n$. 

\begin{figure}[!h]
    \centering
    \includegraphics[width=0.95\columnwidth]{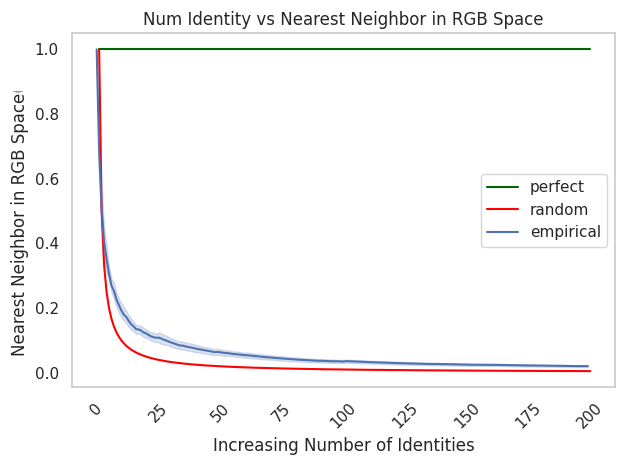}
    \caption{Fraction of nearest neighbor of an image of an identity, when mapped to the RGB space, appears to be nearly random.}
    \label{fig:rgb_cluster}
\end{figure}

To check the importance of color in separating different identities, we randomly select 200 identities from WebFace4M training data, with each identity having a large number of images. 
To select suitable quality images for analysis of the face skin region, we use a BiSeNet semantic segmentation \cite{yu_eccv_2018} to filter out faces where less than 30\% of the face area is classified as skin.
This gives us mostly frontal faces, without too much occlusion from glasses or scalp hair.
To avoid inconsistencies stemming from mouth open/closed and facial expression, we focus on the part of the face above the upper lip, and on the pixels classified as skin (omitting eyebrows and eyes), and calculate the average RGB of skin pixels. For each of the 200 identities, we select the 10 images that have the largest number of such skin pixels. We compute the average skin-pixel RGB value for each of the 10 images from 200 identities, and compute the  fraction of points whose nearest neighbor is from the same identity. The experiments were carried out independently ten times, and the results, including the standard deviation, are shown in Figure \ref{fig:rgb_cluster}.

If color is useful to separate images belonging to different identities in the training set, then the 10 images of each identity should form a compact cluster and the fraction of nearest image from the same identity would be 1. If color is a random clue to separate images belonging to different identities, then for the 10 images of each identity the fraction of nearest image from the same identity would be $1/n$. Both of these scenarios are plotted in Figure \ref{fig:rgb_cluster} along with the actual nearest neighbor results. 

From Figure \ref{fig:rgb_cluster}, we can observe that the RGB values of skin pixels hold more meaningful information than random chance when there are few training identities. However, as the number of identities increases, the mean color value tends toward random information. (RGB mapping of identities with significantly different skin tones can be found in the supplementary material.) This can explain why the network generally does not learn to extract color-oriented features. %
Building upon this finding, we now delve into an investigation of what the network learns when there are only a limited number of filters available in the first convolution layer for training.

\begin{figure*}[!h]
    \centering
    \includegraphics[width=.975\textwidth]{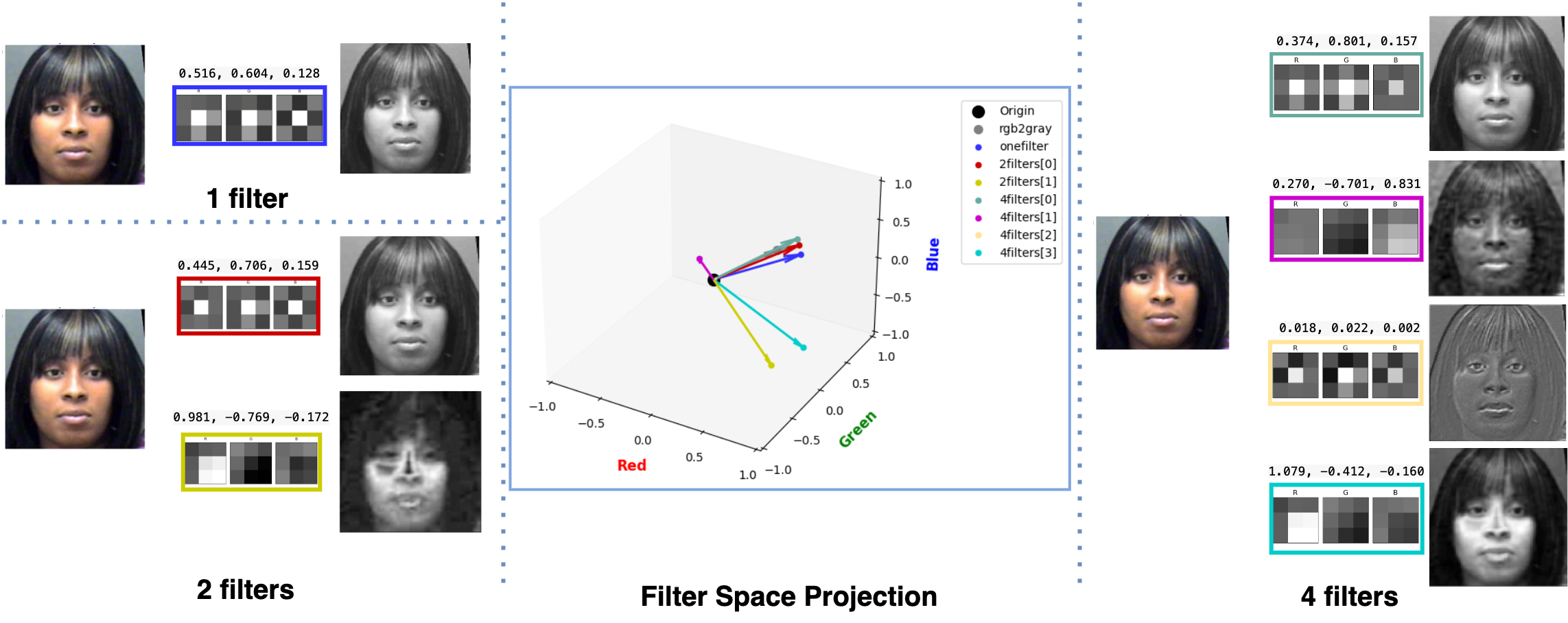}
    \caption{The first and last columns display the activation maps alongside the learned filters for models trained with one, two, and four filters in the first layer. The sum of elements for each filter, as expressed in Equation \ref{kernelsum}, is displayed above each respective filter. The middle column illustrates the vector projection of these learned filters. (Projection vectors betters visualized in color)}
    \vspace{-0.5em}
    \label{fig:filter_forms}
\end{figure*}
\subsection{Training Fewer Filters Learns GrayScale Conversion First}\label{filterforms}

In the previous section, our focus was on determining the importance of color-related information from a training data perspective. In this section, we delve deeper into the model's perspective, specifically examining the features that the model learns in the earlier layers when it has fewer filters than usual to learn. In a typical ResNet architecture, the initial convolution layers are configured with 64 filters. This design implicitly assumes the necessity of 64 unique filters for adequately capturing the variations in a face. The filters in these layers are crucial as they are the sole components directly processing the face image. Beyond these layers, the network operates on  abstract representations derived from the initial layer's output. By deliberately reducing the number of filters in this layer, we can gain insights into the nature of most important information relayed to subsequent layers. Given that grayscale performs similarly to color images, this investigation becomes crucial in understanding the filters the network learns when it has fewer filters to work with compared to the standard training setup. \\

\noindent\textbf{Impact of Reduced Initial Filters on Accuracy.} For this experiment, we chose the ArcFace loss trained using ResNet-50 backbone (See $\S$ \ref{deeperlook} for explanations). We intentionally reduced the number of filter patterns in the first layer to just one, two, and four. 
This allows us to consider what the network learns as the most important one, two or four filters.
However, before we delve into further analysis, it's crucial to check if changing the number of filter patterns in the first layer significantly reduces the network's accuracy. The accuracy results are presented in Table \ref{tab:filters_exp}. The results demonstrate that even when using only one, two, or four filter patterns in the first layer, there is no substantial reduction in the accuracy of the final model.

\setlength\extrarowheight{2pt}
\begin{table}[!h]
\renewcommand{\arraystretch}{1.3}
\caption{Separating the ``chroma" and ``luma" information does not neccesarily change the pattern of results when compared to training in the RGB color space or using grayscale images. 
Key: \textcolor[HTML]{C41E3A}{$p \textcolor[HTML]{C41E3A}{\textless 0.05} $ ; lower accuracy + significant difference}, \textcolor[HTML]{006400}{$p \textgreater 0.05$ ; no significant difference or higher accuracy}
}
\centering
\resizebox{0.9\columnwidth}{!}{%
\begin{tabular}{c|c||ccc}
\hline
\begin{tabular}[c]{@{}c@{}}\diagbox{Test Sets $\downarrow$}{Num. Filters $\rightarrow$ }\end{tabular} & \begin{tabular}[c]{@{}c@{}}64 filters\\ (Baseline)\end{tabular} & 1 filter & 2 filters & 4 filters \\ \hline
LFW      & $99.77 \pm 0.03$  & $99.82$  & $99.80$ & $99.78$  \\
CFP-FP   & $99.03 \pm 0.06$ & $99.01$ & $99.08$ & $98.97$ \\
AGEDB-30 & $97.56 \pm 0.14$ & $97.68$  & $97.78$  & $97.73$ \\
CALFW    & $95.98 \pm 0.08$ & $96.00$ & $96.06$ & $95.98$ \\
CPLFW    & $94.07 \pm 0.10$ & $93.75$ & $94.00$  & $94.23$  \\ \hline\hline
Average  & $97.28 \pm 0.08$ &$97.25$  & $97.34$ & $97.34$ \\ \hline
\end{tabular}
}
\label{tab:filters_exp}
\end{table}

Given the results demonstrating that reducing the number of filters doesn't significantly diminish accuracy, we extend our analysis to entail both a qualitative exploration by visualizing the activation outputs passed to the subsequent layer when the model is trained with fewer filters in the first layer and a quantitative evaluation of how the learned filter projects within the filter space for the learned models. \\

\noindent\textbf{Analysis of Activation Maps.} In Figure \ref{fig:filter_forms}, we observe the visualizations of learned filters in the first column and their corresponding activation maps in the last column. When the model is trained with just one filter, it appears to convert the input RGB image into grayscale before passing it to the next layer. This same conversion pattern is noticeable even when the model is trained with two or four filters in the initial layer. Essentially, one of the learned filters seems to specialize in grayscale conversion, while the remaining filters are dedicated to   tasks such as edge detection or skin detection. \\

\noindent\textbf{Learned Filter Projection in RGB Space.} Now, let's consider representing the weighted average conversion of an RGB image to grayscale in vector form :
\begin{equation}
    rgb2gray(\vec{u}) = 0.299\cdot\vec{r}+0.587\cdot\vec{g}+0.114\cdot\vec{b} 
\end{equation}

This conversion factor is utilized in packages such as OpenCV for  conversion of RGB to grayscale \cite{opencv}. 
When performing convolution, a unit operation involves the element-wise summation across all values in a convolution filter. Thus, we aggregate all the values within the filter for each channel and project the resulting filter conversion vectors into RGB space, as illustrated in the middle column of Fig \ref{fig:filter_forms}. The representation of the projection of the learned filter in the filter space can be expressed as:

\begin{equation}\label{kernelsum}
\small
    proj(\vec{v})= \left(\sum_{i=1}^{m \times n} K_R\right) \cdot\vec{r} + \left(\sum_{i=1}^{m \times n} K_G\right) \cdot\vec{g} + \left(\sum_{i=1}^{m \times n} K_B\right) \cdot\vec{b} 
\end{equation}

\noindent where $proj(\vec{v})$ denotes the projected vector of the learned filter, $K_R$, $K_G$, and $K_B$ represent the filter kernels learned for the red, green, and blue channels, respectively. The filter kernel in the initial layer has dimensions of $3 \times 3$, which means that both `m' and `n' are equal to 3. All these values are presented alongside the visualization of the learned filters in Figure \ref{fig:filter_forms}. Note that visualization of learned filter and projection in the vector space are coded in the same color. A clear pattern emerges when examining the activation maps. Those responsible for grayscale conversion exhibit filter vector representations in RGB space that closely align with RGB to grayscale conversion filters in both magnitude and direction. On the other hand, filter forms designed to extract different features show learned filter vector representations that are distinctly distant from the RGB to grayscale conversion filters, both in terms of both magnitude and direction. These results indicate that the most important feature for the models is  effectively to learn the grayscale conversion operations first, as this is what the model does when it only has one filter to learn.

\subsection{Color vs Gray in the Synthetic Training Dataset} \label{synthetic}
Recent research has highlighted privacy issues concerning the use of real-world datasets for training \cite{ambardi_access_2023,bae_wacv_2023,jain_arxiv_2023,jain_arxiv_2023_zero,kim_cvpr_2023dcface,qiu_cvpr_2021}. Consequently, the generation of synthetic data for facial recognition tasks is on the rise, thanks to improved face image quality from generative methods based on GANs \cite{goodfellow_nips_2014} and DDPM \cite{ho_nips_2020} . Therefore, 
we investigate whether the insights about color versus grayscale apply to a synthetic  training set as well. Further, the synthetic image generation pipeline typically involves uses of significant color and photometric augmentations during training. These methods could lead to problems like uneven color gradients, blurry edges, and other color artifacts in the images, which might adversely affect the network's training. 

To explore whether using color or grayscale has a significant impact in synthetic training datasets, we employed a top-performing synthetic facial training dataset from recent publications \cite{kim_cvpr_2023dcface}. We conducted a similar study on this dataset as we did with the real training dataset and the results are shown in Table \ref{tab:synthetic}. We used ArcFace and AdaFace loss functions with varying backbone sizes for training, using the same configuration file. It's worth noting that fine-tuning the hyperparameters for the synthetic dataset might lead to higher overall accuracy. However, our objective here is not hyperparameter optimization, but rather to understand whether models trained on RGB as opposed to grayscale synthetic datasets perform differently. As long as our configurations for RGB and grayscale training are consistent and result in fairly effective models, our findings should also apply to fully optimized models.

\setlength\extrarowheight{3pt}
\renewcommand{\arraystretch}{1.3}
\begin{table}[!h]
\caption{ Performance comparison of RGB and grayscale synthetic training data for different model depths. Results are highly consistent with training on web-scraped datasets. Differences are observed for shallower networks, but as the model depth increases, the accuracy gap diminishes.
Key: \textcolor[HTML]{C41E3A}{$p \textcolor[HTML]{C41E3A}{\textless 0.05} $ ; lower accuracy + significant difference}, \textcolor[HTML]{006400}{$p \textgreater 0.05$ ; no significant difference or higher accuracy} 
}
\centering
\resizebox{\columnwidth}{!}{
\begin{tabular}{c|c|ccc||ccc}
\hline
          &  Loss $\rightarrow$          &     & ArcFace &         &     & AdaFace &         \\ \hline
Backbone  & \begin{tabular}[c]{@{}c@{}}\diagbox{Test $\downarrow$}{Train $\rightarrow$ }\end{tabular} & RGB & Gray(3)     & p-value & RGB & Gray(3)     & p-value \\ \hline
          & LFW        & 97.88 $\pm$ 0.09     & 97.76       &  \textcolor[HTML]{C41E3A} {\textless 0.05}       & 98.22 $\pm$ 0.06    &  97.78       & \textcolor[HTML]{C41E3A} {\textless 0.05}        \\
          & CFP-FP     & 77.56 $\pm$ 0.42    &  75.78       & \textcolor[HTML]{C41E3A} {\textless 0.05}         & 80.52 $\pm$ 0.32   & 79.35        & \textcolor[HTML]{C41E3A} {\textless 0.05}        \\
ResNet-18 & AGEDB-30   & 88.25 $\pm$ 0.29    &  88.08       & \textcolor[HTML]{006400}{0.26}        & 89.22 $\pm$ 0.21   &  88.93       &  \textcolor[HTML]{C41E3A} {\textless 0.05}       \\
          & CALFW      & 91.14 $\pm$ 0.11    &  90.85       & \textcolor[HTML]{C41E3A} {\textless 0.05}        & 91.28 $\pm$ 0.08    &  91.02       & \textcolor[HTML]{C41E3A} {\textless 0.05}        \\
          & CPLFW      & 79.65 $\pm$ 0.45    & 78.38        & \textcolor[HTML]{C41E3A} {\textless 0.05}        & 81.06 $\pm$ 0.23    &  80.56       &  \textcolor[HTML]{C41E3A} {\textless 0.05}       \\ \cdashline{2-8}
          & Average    & 86.89 $\pm$ 0.15    & 86.17        & \textcolor[HTML]{C41E3A} {\textless 0.05}        & 88.06 $\pm$ 0.04    & 87.53        &  \textcolor[HTML]{C41E3A} {\textless 0.05}       \\ \hline\hline

          & LFW        & 98.02 $\pm$ 0.08    & 97.90        & \textcolor[HTML]{C41E3A} {\textless 0.05}        & 98.38 $\pm$ 0.07    & 98.36        & \textcolor[HTML]{006400}{0.55}        \\
          & CFP-FP     & 80.38 $\pm$ 0.58   & 79.21        & \textcolor[HTML]{C41E3A} {\textless 0.05}        & 85.19 $\pm$ 0.27     & 84.50        & \textcolor[HTML]{C41E3A} {\textless 0.05}        \\
ResNet-50 & AGEDB-30   & 88.33 $\pm$ 0.19   & 88.73        & \textcolor[HTML]{C41E3A} {\textless 0.05}        & 90.25 $\pm$ 0.06    & 90.02 & \textcolor[HTML]{C41E3A} {\textless 0.05}        \\
          & CALFW      & 91.06 $\pm$ 0.06    & 90.90       & \textcolor[HTML]{C41E3A} {\textless 0.05}        & 91.68 $\pm$ 0.20     & 90.78        & \textcolor[HTML]{C41E3A} {\textless 0.05}        \\
          & CPLFW      & 80.49 $\pm$ 0.19   & 79.68        & \textcolor[HTML]{C41E3A} {\textless 0.05}        & 82.82 $\pm$ 0.19    &  82.28       & \textcolor[HTML]{C41E3A} {\textless 0.05}  \\ \cdashline{2-8}
          & Average    & 87.65 $\pm$ 0.15    & 87.29        & \textcolor[HTML]{C41E3A} {\textless 0.05}        & 89.66 $\pm$ 0.06    & 89.19        & \textcolor[HTML]{C41E3A} {\textless 0.05} \\\hline\hline

          & LFW        & 98.18 $\pm$ 0.13    &  97.85       & \textcolor[HTML]{C41E3A} {\textless 0.05}        & 98.58 $\pm$ 0.10    & 98.20        &  \textcolor[HTML]{C41E3A} {\textless 0.05}       \\
          & CFP-FP     & 80.46 $\pm$ 0.51   &  80.14       & \textcolor[HTML]{006400}{0.23}        & 86.59 $\pm$ 0.58    & 85.44        & \textcolor[HTML]{C41E3A} {\textless 0.05}        \\
ResNet-101& AGEDB-30   & 88.85 $\pm$ 0.38   &  89.05       & \textcolor[HTML]{006400}{0.30}        & 90.65 $\pm$ 0.42    & 90.36        & \textcolor[HTML]{006400}{0.20}        \\
          & CALFW      & 91.15 $\pm$ 0.22   & 91.05        & \textcolor[HTML]{006400}{0.37}        & 91.77 $\pm$ 0.13    & 92.03        & \textcolor[HTML]{006400} {\textless 0.05}        \\
          & CPLFW      & 80.62 $\pm$ 0.24  &  80.63       & \textcolor[HTML]{006400}{0.93}        & 83.61 $\pm$ 0.19    & 82.83        & \textcolor[HTML]{C41E3A} {\textless 0.05}        \\\cdashline{2-8}
          & Average    & 87.84 $\pm$ 0.12   & 87.75        & \textcolor[HTML]{006400}{0.17}        & 90.24 $\pm$ 0.18    & 89.77        & \textcolor[HTML]{C41E3A} {\textless 0.05}       \\\hline
\end{tabular}
}
\label{tab:synthetic}
\end{table}

Using grayscale synthetic images for training is consistently worse than using color synthetic images for shallower models like ResNet-18. For slightly deeper models like ResNet-50, the model trained with grayscale synthetic images still performs worse, similar to training with real images. However, for very deep models like ResNet-101, training with color synthetic images is not consistently better or worse than using grayscale synthetic images.

We postulate that although synthetic datasets are improving in their sampling of different images of the same identity, they are not free of aliasing effects like distortions and irregularities. Given that shallower models rely on low-level features, these artifacts, combined with the inability of shallower models to represent intricate features, could contribute to the accuracy disparities between synthetic color and grayscale training datasets for shallower models.

\setlength\extrarowheight{2pt}
\begin{table*}[h!]
\renewcommand{\arraystretch}{1.3}
\caption{
\textbf{Training with Additional Data using the Freed Disk Space can improve Grayscale accuracy compared to RGB}. 1:1 Verification accuracy (\%) when trained and evaluated all in single-channel grayscale versus trained and evaluated in RGB is essentially the same. Improving accuracy is possible by using additional data in the freed-up disk space. RGB  and Gray  represents the color-cleaned WebFace4M in RGB and one-channel Grayscale format. \textbf{Gray+ } represents the data in Gray + additional one-channel grayscale data pooled from WebFace12M to make use of emptied disk space. Key: \textcolor[HTML]{0000FF}{$p \textless 0.05 $ ; higher accuracy + significant difference}, \textcolor[HTML]{C41E3A}{$p \textless 0.05 $ ; lower accuracy + significant difference}, \textcolor[HTML]{006400}{$p \textgreater 0.05$ ; no significant difference}
}
\centering
\resizebox{\textwidth}{!}{%
\begin{tabular}{c|c|c|ccccc||ccccc}
\hline
Backbone& &Loss $\rightarrow$    &       &       & ArcFace &       &         &       &       & AdaFace &       &         \\ \cline{3-13}
&Protocol $\downarrow$&\begin{tabular}[c]{@{}l@{}}\diagbox{Test $\downarrow$}{Train $\rightarrow$ }\end{tabular}  & RGB (3)  & Gray (1)  & p-value & Gray+  & p-value & RGB (3) & Gray (1) & p-value & Gray+ (1) & p-value \\ \hline
& &LFW      & 99.60 $\pm$ 0.08  & 99.55 & \textcolor[HTML]{006400} {0.23}  & 99.65 & \textcolor[HTML]{006400} {0.23} & 99.57 $\pm$ 0.03  & 99.63  & \textcolor[HTML]{0000FF} {\textless 0.05} & 99.60 & \textcolor[HTML]{006400} {0.09}  \\
& &CFP-FP   & 97.71 $\pm$ 0.18  & 97.45  & \textcolor[HTML]{C41E3A} {\textless 0.05}  & 97.57 & \textcolor[HTML]{006400} {0.16}  & 97.27 $\pm$ 0.11 & 96.95 & \textcolor[HTML]{C41E3A} {\textless 0.05} &97.25 & \textcolor[HTML]{006400} {0.70}  \\
&1:1 Verificaton Accuracy(\%) &AGEDB-30 & 96.63 $\pm$ 0.19  & 96.53 & \textcolor[HTML]{006400} {0.30}  & 96.83  & \textcolor[HTML]{006400} {0.07}  & 96.24 $\pm$ 0.18 & 96.10 & \textcolor[HTML]{006400} { 0.16 } & 96.06  & \textcolor[HTML]{006400} { 0.09 }  \\
&&CALFW    & 95.58 $\pm$ 0.08 & 95.70 & \textcolor[HTML]{C41E3A} {\textless 0.05}  & 95.72 & \textcolor[HTML]{C41E3A} {\textless 0.05} & 95.35 $\pm$ 0.20 & 95.30 & \textcolor[HTML]{006400} { 0.60 }     & 95.12 & \textcolor[HTML]{006400} { 0.07 }\\
ResNet-18&&CPLFW    & 92.39 $\pm$ 0.17  & 91.98 & \textcolor[HTML]{C41E3A} {\textless 0.05}   & 91.92 & \textcolor[HTML]{C41E3A} {\textless 0.05}  &  91.95 $\pm$ 0.07 & 91.73 &  \textcolor[HTML]{C41E3A} {\textless 0.05}   & 91.80 & \textcolor[HTML]{C41E3A} {\textless 0.05}   \\ \cdashline{3-13}
&&Average  & 96.38 $\pm$ 0.07  & 96.24 & \textcolor[HTML]{C41E3A} {\textless 0.05} & 96.34 & \textcolor[HTML]{006400} {0.27} & 96.08 $\pm$ 0.09 &  95.94 & \textcolor[HTML]{C41E3A} {\textless 0.05}   & 95.97  & \textcolor[HTML]{006400} { 0.05 }  \\ \cline{2-13}
&&IJB-B  & 93.12 $\pm$ 0.07 & 92.70 & \textcolor[HTML]{C41E3A} {\textless 0.05}  & 93.13  & \textcolor[HTML]{006400} {0.76} &  93.06 $\pm$ 0.09 & 92.35 & \textcolor[HTML]{C41E3A} {\textless 0.05}  & 92.41  & \textcolor[HTML]{C41E3A} {\textless 0.05}  \\ 
&TAR@FAR = 0.01\%&IJB-C  & 95.16 $\pm$ 0.02 & 94.77 & \textcolor[HTML]{C41E3A} {\textless 0.05}  & 95.09
& \textcolor[HTML]{C41E3A} {\textless 0.05} &  94.95 $\pm$ 0.05 & 94.56 & \textcolor[HTML]{C41E3A} {\textless 0.05}  & 94.49 & \textcolor[HTML]{C41E3A} {\textless 0.05} \\ \hline\hline

&&LFW      & 99.77 $\pm$ 0.03 & 99.78 & 0.74  & 99.83 & \textcolor[HTML]{0000FF} {\textless 0.05} &  99.77 $\pm$ 0.02 & 99.78 & \textcolor[HTML]{006400}{0.32} & 99.77 & \textcolor[HTML]{006400}{1.0}  \\
&&CFP-FP   & 99.03 $\pm$ 0.06 & 99.11 & 0.05 & 99.21 & \textcolor[HTML]{0000FF} {\textless 0.05} & 98.76 $\pm$ 0.04 & 98.75  & \textcolor[HTML]{006400}{0.60} & 98.90  & \textcolor[HTML]{0000FF} {\textless 0.05} \\
&1:1 Verificaton Accuracy(\%) &AGEDB-30 &  $97.56\pm0.14$ & 97.85 & \textcolor[HTML]{0000FF} {\textless 0.05}  & 97.88 & \textcolor[HTML]{0000FF} {\textless 0.05}  &  97.48 $\pm$ 0.05 & 97.33 & \textcolor[HTML]{C41E3A} {\textless 0.05} & 97.60 & \textcolor[HTML]{0000FF} {\textless 0.05}  \\
&&CALFW    & 95.98 $\pm$ 0.08 & 96.10 & \textcolor[HTML]{0000FF} {\textless 0.05} & 96.08 & \textcolor[HTML]{0000FF} {\textless 0.05} & 95.96 $\pm$ 0.06 & 95.98 & \textcolor[HTML]{006400}{0.49} & 95.85 & \textcolor[HTML]{C41E3A} {\textless 0.05} \\
ResNet-50&&CPLFW    &  94.07 $\pm$ 0.10 & 93.85 & \textcolor[HTML]{006400}{0.35}  & 94.10 & \textcolor[HTML]{006400}{0.53}  &  93.87 $\pm$ 0.10 & 93.76  & \textcolor[HTML]{006400}{0.07}   & 93.65 & \textcolor[HTML]{C41E3A} {\textless 0.05} \\ \cdashline{3-13}
&&Average  & 97.28 $\pm$ 0.08 & 97.34 & \textcolor[HTML]{0000FF} {\textless 0.05} & 97.42 & \textcolor[HTML]{0000FF} {\textless 0.05} & 97.18 $\pm$ 0.04 & 97.12 & \textcolor[HTML]{C41E3A} {\textless 0.05} & 97.16 & \textcolor[HTML]{006400}{0.33}   \\ \cline{2-13}
&TAR@FAR = 0.01\%&IJB-B  & 95.16 $\pm$ 0.09 & 95.19 & \textcolor[HTML]{006400}{0.49}  & 95.48 & \textcolor[HTML]{0000FF} {\textless 0.05} & 95.26 $\pm$ 0.03 & 95.03 & \textcolor[HTML]{C41E3A} {\textless 0.05}  &95.27  & \textcolor[HTML]{006400}{0.49}   \\ 
&&IJB-C  & $96.88 \pm 0.08$ & $96.86$ & \textcolor[HTML]{006400}{0.60}  & 97.05 & \textcolor[HTML]{0000FF} {\textless 0.05} & 96.77 $\pm$ 0.08 & 96.59 & \textcolor[HTML]{C41E3A} {\textless 0.05} &96.79  & \textcolor[HTML]{006400}{0.60}  \\ \hline\hline

&&LFW      & 99.78 $\pm$ 0.03  & 99.85 & \textcolor[HTML]{0000FF} {\textless 0.05}  & 99.83 & \textcolor[HTML]{0000FF} {\textless 0.05} & 99.80 $\pm$ 0.03 & 99.73  & \textcolor[HTML]{C41E3A} {\textless 0.05} & 99.85 & \textcolor[HTML]{0000FF} {\textless 0.05}   \\
&&CFP-FP   & 99.18 $\pm$ 0.05  & 99.28 & \textcolor[HTML]{0000FF} {\textless 0.05} & 99.33 & \textcolor[HTML]{0000FF} {\textless 0.05} & 99.07 $\pm$ 0.09 & 99.11  & \textcolor[HTML]{006400}{0.38} & 99.09  & \textcolor[HTML]{006400}{0.64} \\
&1:1 Verificaton Accuracy(\%) &AGEDB-30 & 97.99 $\pm$ 0.10  & 97.80 & \textcolor[HTML]{C41E3A} {\textless 0.05}  & 98.15 & \textcolor[HTML]{0000FF} {\textless 0.05}  & 97.74 $\pm$ 0.04 & 97.95 & \textcolor[HTML]{0000FF} {\textless 0.05}  & 97.77 & \textcolor[HTML]{006400} {0.17}  \\
&&CALFW    & 96.06 $\pm$ 0.07  & 95.98 & \textcolor[HTML]{006400}{0.06} & 96.15 & \textcolor[HTML]{0000FF} { \textless 0.05} & 96.02 $\pm$ 0.12 & 95.92  & \textcolor[HTML]{006400}{0.13}  & 96.00 & \textcolor[HTML]{006400}{0.73} \\
ResNet-101&&CPLFW    & 94.42 $\pm$ 0.09 & 94.28 & \textcolor[HTML]{C41E3A} {\textless 0.05}  & 94.35 & \textcolor[HTML]{006400}{0.16}  & 94.34 $\pm$ 0.09  & 94.16 & \textcolor[HTML]{C41E3A} {\textless 0.05}   & 94.42 & \textcolor[HTML]{006400}{0.12}  \\ \cdashline{3-13}
&&Average  & 97.45 $\pm$ 0.11 & 97.44 & \textcolor[HTML]{006400}{0.84}  & 97.56 & \textcolor[HTML]{006400}{0.08} & 97.40 $\pm$ 0.06 & 97.37 & \textcolor[HTML]{006400}{0.32}  & 97.43 & \textcolor[HTML]{006400}{0.32}  \\ \cline{2-13}
&TAR@FAR= 0.01\%&IJB-B  & 95.71 $\pm$ 0.08 & 95.66 & \textcolor[HTML]{006400}{0.23}  & 95.97 & \textcolor[HTML]{0000FF} {\textless 0.05} & 95.72 $\pm$ 0.11 & 95.60 & \textcolor[HTML]{006400}{0.08} & 95.73  & \textcolor[HTML]{006400}{0.85}  \\ 
& &IJB-C  & 97.28 $\pm$ 0.06 & 97.26 & \textcolor[HTML]{006400}{0.49} & 97.38 & \textcolor[HTML]{0000FF} {\textless 0.05} & 97.13 $\pm$ 0.09 & 97.02 & \textcolor[HTML]{006400}{0.06}  & 97.06  & \textcolor[HTML]{006400}{0.16}  \\  \hline

\end{tabular}
}
\vspace{0.2cm}

{\scriptsize $\Delta$  Difference of the accuracies computed as RGB - Gray/+ }
\label{tab:one-channel}
\end{table*}

\section{Computing used more effectively in grayscale}\label{onechanneltraining}
In the analyses in earlier section, the grayscale images were in a 3-channel format with R=G=B.
The results suggest that color has no consistent advantage over grayscale for deeper face recognition networks.
If this is the case, then training and testing a network using only single-channel grayscale images should give essentially the same accuracy as training and testing on RGB color images. However, native grayscale images use less disk space than RGB. If we use the freed-up disk space with additional data, can it enhance the model's accuracy? Additionally, does employing grayscale make the training process more efficient? This section explores these questions.

We modify the ResNet backbone \cite{he_cvpr_2016,Deng_CVPR_2019} so that the first convolutional layer processes a single-channel image rather than a three-channel image.
Due to modifications in the backbone, we also utilize single-channel grayscale test images instead of color images for testing, as described in $\S$ \ref{grayscaleimages}.
The size of the first convolution block is changed from \textit{$64\times3\times3\times3$} to \textit{$64\times1 \times3\times3$}. Following that, we train the modified backbone using: a) single-channel grayscale images from the color-cleaned WebFace4M subset, and b) the dataset from (a) combined with additional data from WebFace12M to fill the disk space previously occupied by RGB images, now converted to grayscale. Subsequently, both trained networks are evaluated on single-channel grayscale images.
The accuracy of this fully grayscale face recognition and training is compared to the accuracy of the corresponding fully color network.

\subsection{Accuracy and Efficiency with Single-Channel GrayScale}
Results of this experiment for varying ResNet backbone sizes, with ArcFace and AdaFace loss, are presented in Table \ref{tab:one-channel}. The results of the single-channel grayscale training show a similar pattern to that of three-channel grayscale images. For both loss functions, the shallower ResNet-18 model consistently exhibits lower performance than the model trained on color images. For a deep model like ResNet-50, the grayscale model trained with ArcFace loss shows no consistent significant difference compared to the model trained with color images. However, the AdaFace loss consistently performs worse for the grayscale model. Nonetheless, the accuracy gap diminishes compared to the shallower ResNet-18 model. For deeper models like ResNet-101, for both loss functions and across standard benchmark and more challenging IJB testing suite, there is no consistent difference between the model trained with color or grayscale images.

But, color images require more storage space in disk. For instance, the RGB version of ``color-cleaned" WebFace4M training set, occupies 96 GB of disk storage, while the grayscale version takes up 67 GB. The storage of images in disk thus, is reduced to about 2/3 on average in going to single channel grayscale.\footnote{Additional disk space is allocated to store header files that hold essential image metadata including details like dimensions, color mode, compression method, and other necessary information for accurate image interpretation and display. 
As a result, the occupied disk space is not precisely one-third of the total disk space for RGB images.} 
This reduction in disk space could be beneficial for training deep CNN face networks. During training, GPU memory comprises of the model itself, mini-batch for training with some additional overheads. Large training datasets exceed GPU memory capacity, requiring continuous transfer of mini-batches from the disk to the GPU for training. Grayscale images have a smaller memory footprint (1/3 of bytes to represent the tensor) than RGB, reducing data transfer volume between CPU and GPU during training.
Though for a single forward pass, the computational and training efficiency doesn't improve significantly,  overall training is improved due to less data transfer between CPU and GPU. For example, with a cluster of 4 Titan-Xp GPUs hosted on Intel(R) Xeon(R) CPU E5-2650 v4 @ 2.20GHz system, we observed approximately 1.2K GBs of data transferred during RGB training and about 0.48K GBs for single-channel grayscale training. This led to an approximately 20\% improvement in system CPU time used. 
The exact savings will vary between systems.
\vspace{-0.5em}

\subsection{Improved Accuracy by Utilizing Freed-Up Disk Space}
Opting for grayscale images to train and test face recognition network delivers comparable performance to color images, while consuming less disk space, which frees up additional disk storage space. To make the most of the available free disk space, we use additional data from WebFace12M, which shares similar characteristics with WebFace4M. The added data is selected randomly to maintain a consistent number of images per identity as ``color-cleaned" WebFace4M. As a result, the total disk space now equals approximately 96 GB, with 282K identities and 5.5M images.  Results for this experiment are in Table \ref{tab:one-channel}.

In the case of ArcFace, using extra data during training (Gray+) boosts the performance for all model depths. With this accuracy boost, the shallower ResNet-18 model now has comparable performance to the model trained on color data, while the deeper ResNet-50 and ResNet-101 models significantly outperform the color model on both standard benchmarks and the more challenging IJBB and IJBC suite. In the case of AdaFace, using extra data during training (Gray+) boosts the performance across all model depths, though the effect is not as pronounced as with ArcFace. For shallower models, the increase in accuracy makes their performance comparable to the color model on standard benchmarks, but the accuracy does not improve significantly for the IJB testing suite. For ResNet-50, the accuracy boost brings the performance on both standard benchmarks and the IJB suite to a level comparable to the color model, whereas previously, it was significantly lower with gray data. For even deeper ResNet-101 models, the accuracy with gray is already within the statistical bounds of the color model. With additional data, the accuracy appears to increase, but this increase does not make the accuracy statistically significantly higher than the baseline model trained on color, as indicated by the p-value shown in Table \ref{tab:one-channel}.

Overall, leveraging the additional data from the freed-up disk space can improve the accuracy of grayscale training across varying model depths for multiple loss functions and learning paradigms.

\section{Conclusions}\label{conclusions}
\noindent\textbf{SoTA deep CNN face matchers do not ``see'' color.}
When the SoTA deep CNN is trained on only grayscale training images (stored in three-channel format) and evaluated on RGB color images, essentially the same accuracy is achieved as when the deep CNN is trained on RGB color images -- the network learns as much from grayscale training data as from color training data, for the purpose of matching color face images.
When the deep CNN is trained on RGB color images, its accuracy on a grayscale or an RGB version of a test set is essentially the same - what the network learns from color training images does not make it any better at matching color versions of test images.

\noindent\textbf{There is no magic color space.} The prevalence of datasets in RGB color space and the default training conducted in this space are well-established. 
Different color spaces may effectively separate chroma and luma information, but our results with HSV  show that changing the color space does not change the basic results.
In fact, the results make it clear that the network relies on the V channel more than the channels with color information.
Our results are consistent with Albiol et al's finding that there is an equivalent optimal skin detector for every color space \cite{Albiol}.

\noindent\textbf{Computing resources used unproductively processing color.}
A deep CNN face matcher trained on single-channel grayscale images, and matching single-channel grayscale images, achieves essentially the same accuracy as a network working with RGB color images.
But the single-channel grayscale images use 1/3 the memory of the RGB images.
And the early convolutional layer of the grayscale network has 1/3 the weights of the color network.

\noindent\textbf{Conditions specific to deep CNN face matching.}
It is known that the color is important for some general object detection tasks solved by deep CNNs \cite{de_2021_ICCV, singh_2020_arxiv}.
Our results here are not in conflict with these studies.
Deep CNN face matching is a specialized task.
Matchers are trained to recognize (categorize) persons from in-the-wild, web-scraped images, and color is not consistent across images of a person in this context.
It is possible that a very tightly controlled face matching application, with all images always acquired in the same lighting and with consistent background, could result in color being more useful.

\noindent\textbf{The role of color in demographic accuracy differences.} It is acknowledged that face recognition accuracy varies across demographic groups ~\cite{albiero_gender, bhatta_wacvw_2023,bhatta_wacvw_2024,cook_tbiom_2019,cook2023demographic,dhar_iccv_2021,drozdowski_tts_2020,fu_ijcb_2022,huang_cvpr_2023,huber_arxiv_2023,jain_arxiv_2023,kolla_wacv_2023,kotwal_wacvw_2024,krishnapriya_tifs_2020,krishnapriya_wacvw_2022,liang_iccv_2023,maluleke_eccv_2022, ozturk_arxiv_2023,perera_ijcb_2023,terhorst_tts_2021,villalobos_ieeeaccess_2022,wu_cvprw_2023, wu_cvpr_2023, wu_bmvc_2023,yucer_arxiv_2023}.
Discussion of this topic often mentions skin tone or skin color.
Our results suggest that the accuracy differences are not specific to using images with color content for deep SOTA face recognition networks.
The same differences can be observed when processing grayscale images rather than color.

\ifCLASSOPTIONcaptionsoff
  \newpage
\fi

\bibliographystyle{IEEEtran.bst}
\bibliography{references}

\begin{IEEEbiography}[{\includegraphics[width=1in,height=1.25in,clip,keepaspectratio]{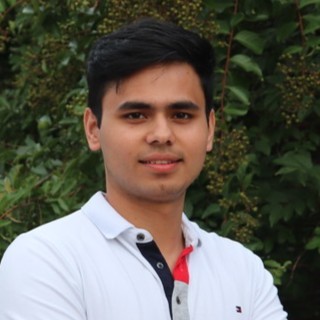}}]{Aman Bhatta}
is a Ph.D. student at the University of Notre Dame. He received the B.S. degree in Mechanical Engineering from the University of Mississippi, in 2021. He is currently pursuing a Ph.D. degree in Computer Science and Engineering at the University of Notre Dame. His research interests include biometrics, computer vision, and machine learning.
\end{IEEEbiography}

\begin{IEEEbiography}[{\includegraphics[width=1in,height=1.25in,clip,keepaspectratio]{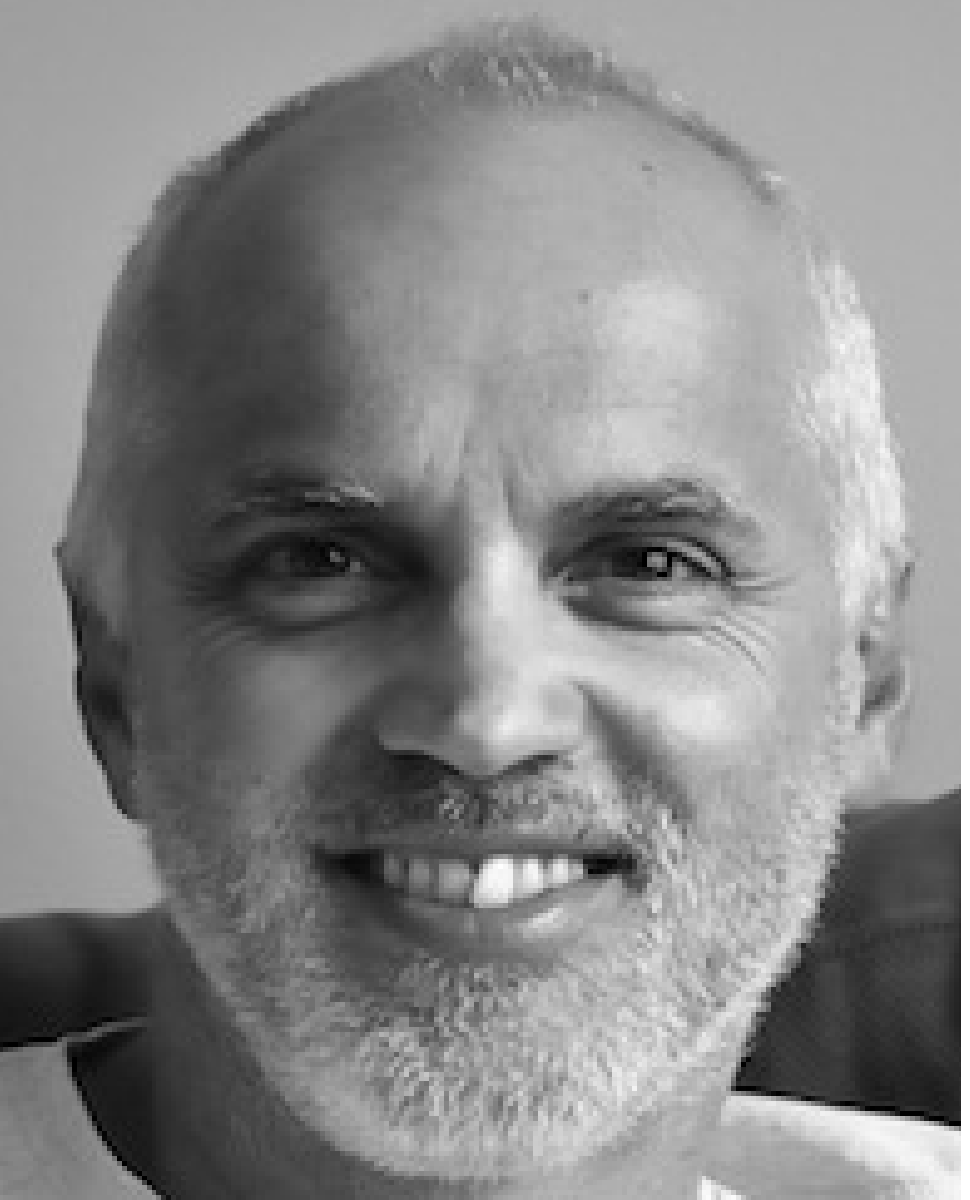}}]{Domingo Mery (Member, IEEE)}
received
the M.Sc. degree in electrical engineering from
the Technical University of Karlsruhe, in 1992,
and the Ph.D. degree (Hons.) from the Technical
University of Berlin, in 2000. In 2001, he worked
as an Associate Researcher with the Department of
Computer Engineering, Universidad de Santiago,
Chile. In 2014, he was a Visiting Professor with
the University of Notre Dame. He is currently a
Full Professor with the Department of Computer
Science, Pontificia Universidad Católica de Chile, where he served as the
Chair, from 2005 to 2009, and the Director of Research and Innovation
with the School of Engineering, from 2015 to 2018. His research interests
include image processing, computer vision applications, and biometrics.
He received the Ron Halmshaw Award, in 2005, 2012, and 2017, and the
John Green Award, in 2013, from the British Institute of Non-Destrsssuctive
Testing, which was established to recognize the best papers published in
the Insight Journal on Industrial Radiography. He received the Best Paper
Award at the International Workshop on Biometrics in conjunction with the
European Conference on Computer Vision (ECCV 2014). He is currently
serving as an Associate Editor for
IEEE Transactions on Information Forensics AND Security, and IEEE
Transactions on Biometrics, Behavior, AND Identity Science.
\end{IEEEbiography}

\begin{IEEEbiography}[{\includegraphics[width=1in,height=1.25in,clip,keepaspectratio]{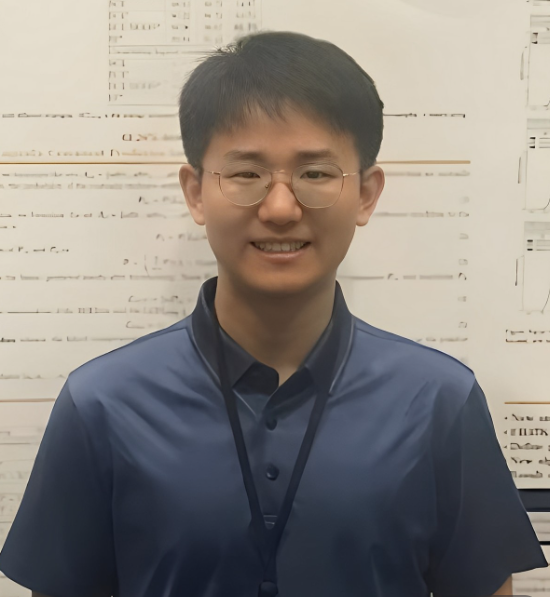}}]{Haiyu Wu} is a Ph.D. student at the University of Notre Dame. He received his B.S. in Electrical Engineering Computer Engineering Emphasis from Northern Arizona University, in 2020. He is currently pursuing a Ph.D. degree in Computer Engieering at the University of Notre Dame. His research interests include improving the ethical responsibility of face recognition algorithms, computer vision, and machine learning.
\end{IEEEbiography}

\begin{IEEEbiography}[{\includegraphics[width=1in,height=1.25in,clip,keepaspectratio]{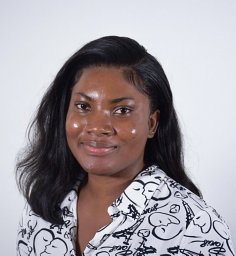}}]{Joyce Annan}
is a Ph.D. student in Computer Science at the Florida Institute of Technology, where she also serves as a graduate research assistant at the L3Harris Institute for Assured Information Identity Laboratory. Her research interests include biometrics, computer vision, data science, and deep learning. Joyce holds an undergraduate degree in Information Technology from the University of Ghana and a master's in Artificial Intelligence and Robotics from the University of Hertfordshire, UK.
\end{IEEEbiography}

\begin{IEEEbiography}[{\includegraphics[width=1in,height=1.25in,clip,keepaspectratio]{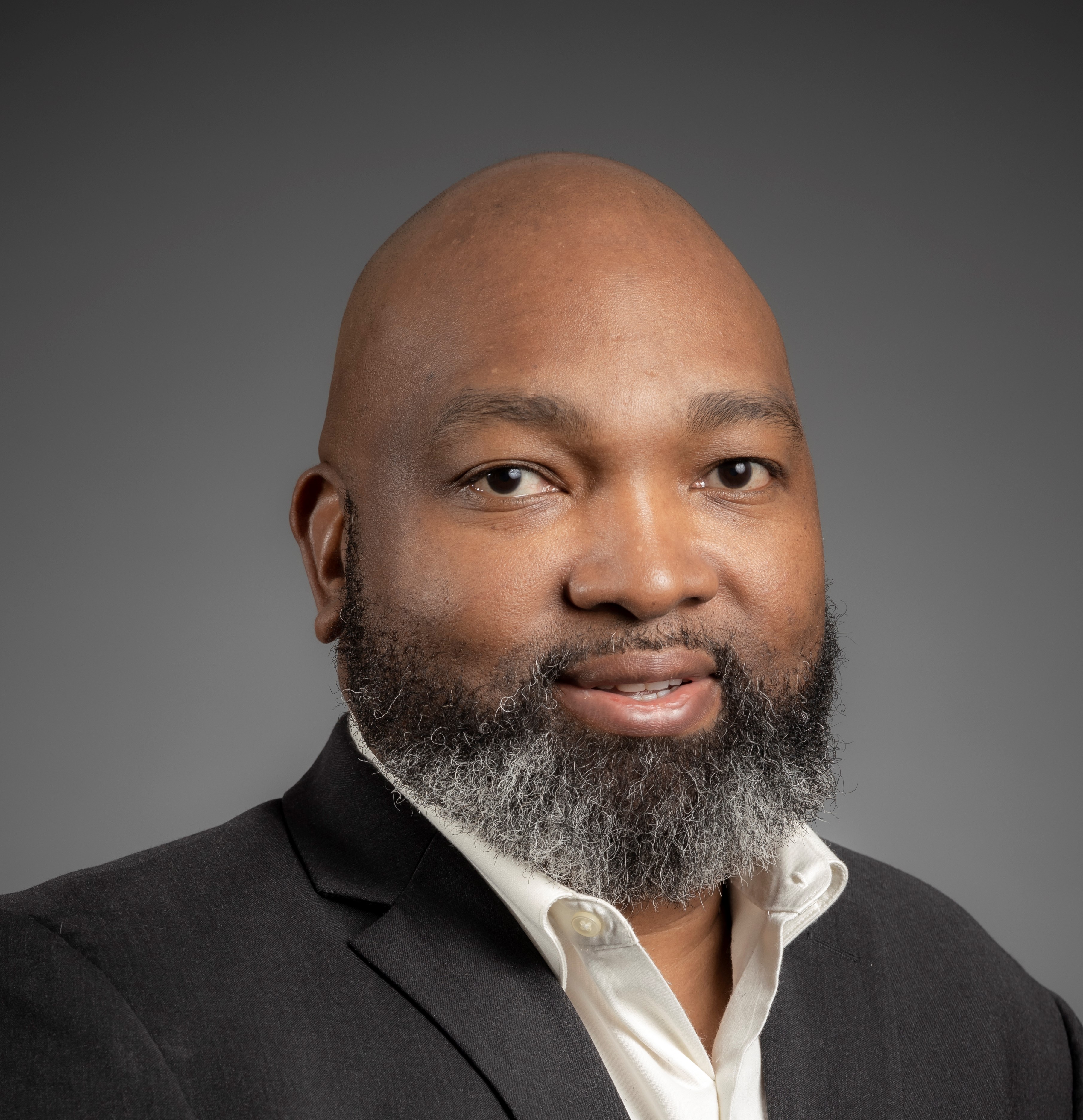}}]{Michael C. King (Senior Member, IEEE)}
joined Florida Institute of Technology’s Harris Institute for Assured Information as a Research Scientist in 2015 and holds a joint appointment as Associate Professor of Computer Engineering and Sciences. Prior to joining academia, Dr. King served for more than 10 years as a scientific research/program management professional in the United States Intelligence Community. While in government, Dr. King created, directed, and managed research portfolios covering a broad range of topics related to biometrics and identity to include: advanced exploitation algorithm development, advanced sensors and acquisition systems, and computational imaging. He crafted and led the Intelligence Advanced Research Projects Activity’s (IARPA) Biometric Exploitation Science and Technology (BEST) Program to transition technology deliverables successfully to several Government organizations.
\end{IEEEbiography}

\begin{IEEEbiography}[{\includegraphics[width=1in,height=1.25in,clip,keepaspectratio]{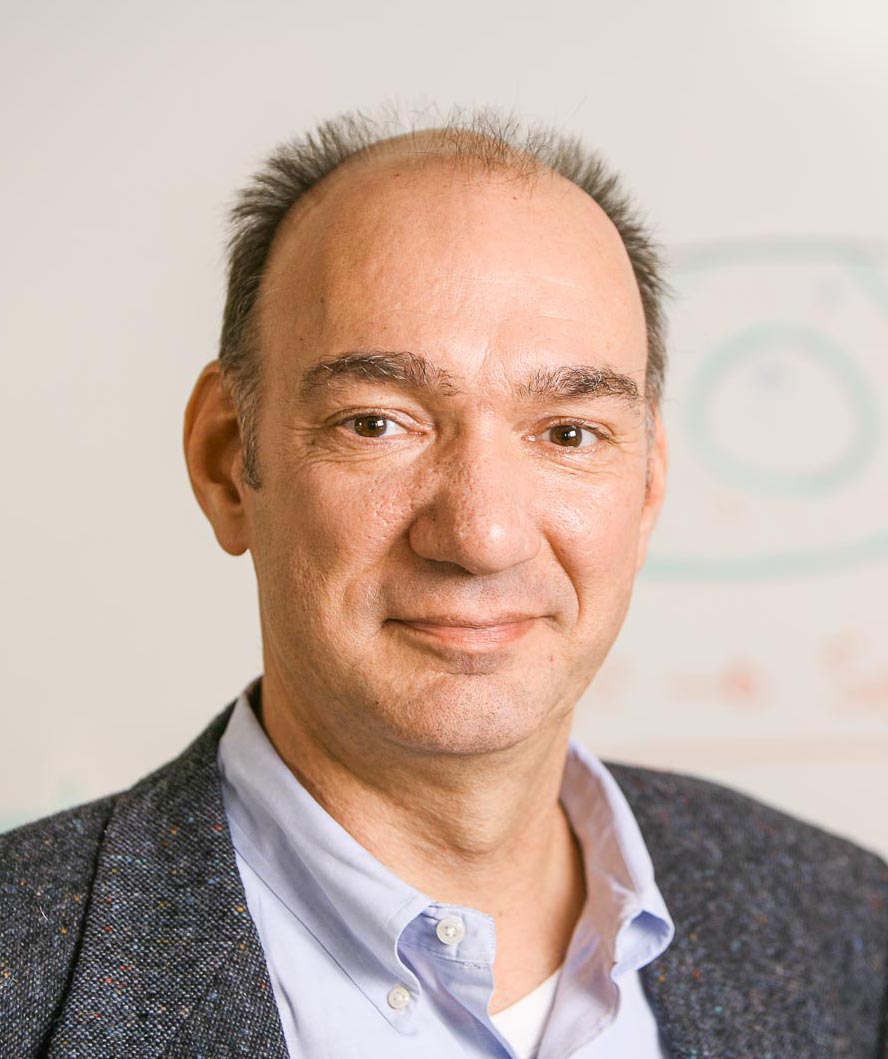}}]{Kevin W. Bowyer (Fellow, IEEE)}
is the Schubmehl-Prein Family Professor of Computer Science and Engineering at the University of Notre Dame, and also serves as Director of International Summer Engineering Programs for the Notre Dame College of Engineering. In 2019, Professor Bowyer was elected as a Fellow of the American Association for the Advancement of Science. Professor Bowyer is also a Fellow of the IEEE and of the IAPR, received a Technical Achievement Award from the IEEE Computer Society, with the citation ``for pioneering contributions to the science and engineering of biometrics''. He previously served as Editor-in-Chief of the IEEE Transactions on Pattern Analysis and Machine Intelligence and IEEE Transactions on Biometrics, Behavior and Identity Science.
\end{IEEEbiography}

\end{document}